%% file: acl_latex.tex
\documentclass[11pt]{article}

\usepackage[preprint]{acl}
\usepackage{times}
\usepackage{latexsym}
\usepackage[T1]{fontenc}
\usepackage[utf8]{inputenc}
\usepackage{microtype}
\usepackage{inconsolata}
\usepackage{graphicx}
\usepackage{hyperref}
\usepackage{url}
\usepackage{xurl}
\usepackage{booktabs}
\usepackage{amsmath}
\usepackage{enumitem}
\usepackage{amssymb}
\usepackage{placeins}
\usepackage{subcaption}
\usepackage{makecell}
\usepackage{siunitx}
\usepackage{multirow}

\usepackage{tikz}
\usepackage{xcolor}
\usetikzlibrary{positioning,calc}

\DeclareMathOperator*{\argmax}{arg\,max}

\newcommand{\divLM}{\textsc{DivLM}}
\newcommand{\divLMshort}{\textsc{DLM}}
\newcommand{\grpocos}{\textsc{\divLMshort$_{\text{cos}}$}}

\makeatletter
\g@addto@macro\normalsize{%
    \setlength{\abovedisplayskip}{2.5pt}
    \setlength{\belowdisplayskip}{2.5pt}
    \setlength{\abovedisplayshortskip}{2.5pt}
    \setlength{\belowdisplayshortskip}{2.5pt}
}
\makeatother

\title{Improving Diversity in LLM Short Story Generation}

\author{Zahra Solati Dehkordi {\normalfont and} Vasileios Lampos\\[0.4em]
  Centre for Artificial Intelligence\\
  Department of Computer Science\\
  University College London, UK\\[0.1em]
  \{\texttt{zahra.solati.24},\,\texttt{v.lampos}\}\texttt{@ucl.ac.uk}}

\begin{document}
\maketitle
\begin{abstract}
Large language models (LLMs) can generate accurate responses, but these are void of diversity. We attempt to address this for the task of creative short story generation. Drawing on established writing conventions and known LLM limitations, we target variation in genre, tone, style, and named entities. To promote diversity across these dimensions, we introduce \divLM, an LLM post-training framework consisting of two phases. First, we perform continued pre-training on a creative writing corpus and restore instruction-following capabilities using weight residuals. We then apply reinforcement learning with a custom, composite reward function that jointly maximizes diversity across the targeted narrative dimensions while maintaining response quality. Our empirical results on two LLM families show that \divLM~increases diversity metrics by more than 9\% on average compared to alternative approaches, while preserving instruction following, overall response quality, and similarity to human outputs.\footnote{Accepted by EACL 2027 (Findings).}

{
\small
\textbf{Source code:} \href{https://github.com/ZahraSolatiDehkordi/DivLM}{\nolinkurl{github.com/ZahraSolatiDehkordi/DivLM}}
\vspace{0.2in}
}
\end{abstract}

\section{Introduction}
Large language models (LLMs) have significantly improved in their ability to follow instructions and produce accurate and helpful responses. These advances have led to a growing interest in using LLMs for creative writing and storytelling~\citep{pei-etal-2024-swag, venkatraman-etal-2025-collabstory, wang-etal-2025-generating}. However, prior studies show that while LLM assistance can increase the perceived creativity of individual human-authored texts, it reduces the collective diversity of content produced by groups of writers~\citep{doshi-hauser-2024-generative, padmakumar2024diversity}. This concern is consistent with evidence that LLMs often exhibit limited diversity across repeated generations from the same or similar prompts~\citep{li-etal-2025-jointly, chakrabarty2024artifice, chu-etal-2025-exploring, yun-etal-2025-price, wang-etal-2025-multilingual, xu2025echoes}. This low diversity is largely attributed to the post-training process, where LLMs are trained using methods such as instruction tuning and reinforcement learning from human feedback (RLHF) to increase output quality and alignment. This can narrow a model's output distribution toward high-reward or typical responses~\citep{padmakumar2024diversity, yun-etal-2025-price, kirk2024understanding}. Such distributional narrowing can lead to mode collapse, where a model repeatedly generates outputs from a limited subset of plausible responses~\citep{hamilton-2024-detecting}. This limitation is particularly important in creative writing workflows, where users often rely on LLMs to explore a space of possible ideas.

\begin{figure*}[t]
\centering
  \includegraphics[width=\linewidth]{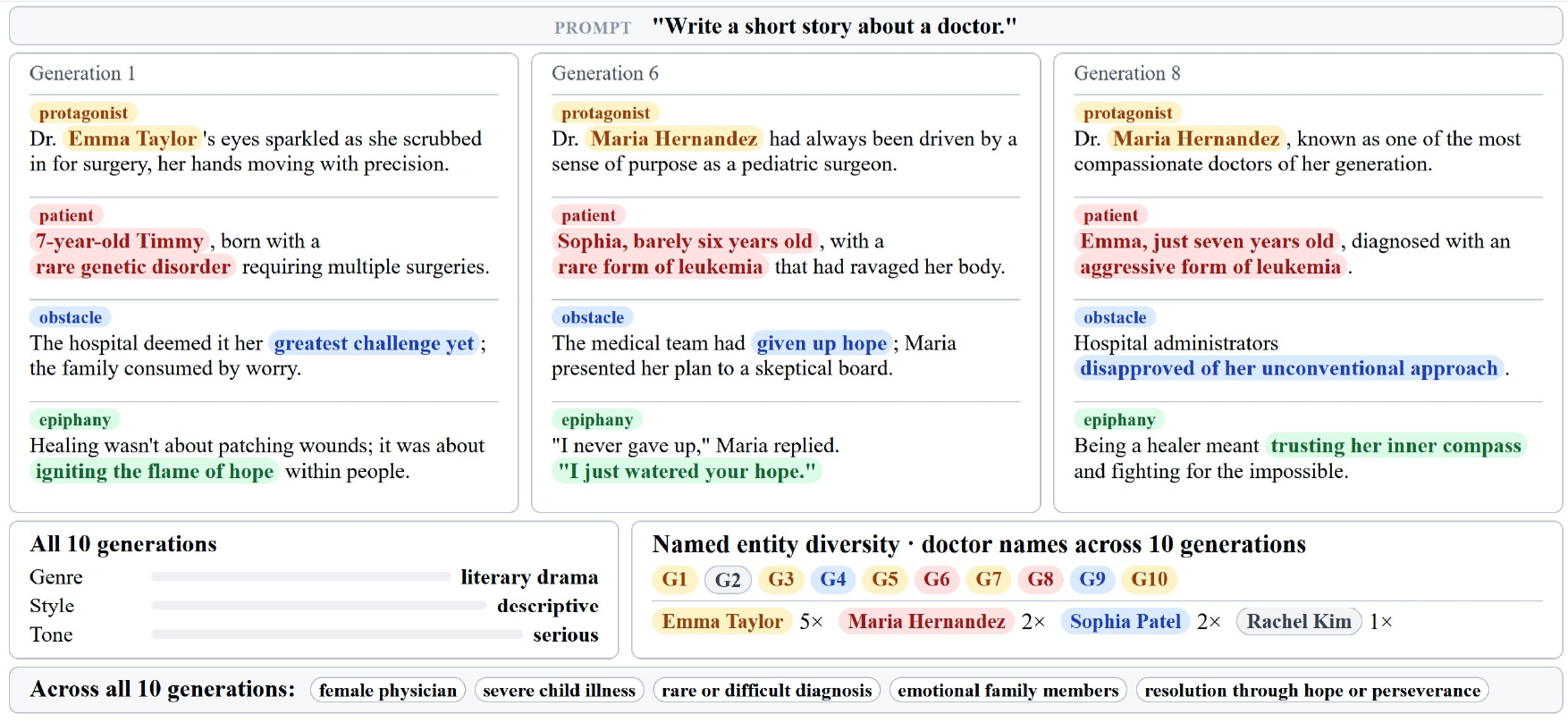}
\caption{Diversity absence (mode collapse) for \texttt{Llama-3.1-8B-Instruct} across generations for the same prompt.}
\label{fig:story-diversity-collapse}
\end{figure*}

A common way to encourage more varied outputs is to increase sampling stochasticity through decoding methods such as temperature scaling. However, prior work has found that such methods do not necessarily improve diversity while coherence or quality drops~\citep{peeperkorn-etal-2024-temperature, chu-etal-2025-exploring}. Other existing approaches include prompt-based~\cite{wang-etal-2025-multilingual,chu-etal-2025-exploring,zhang2025verbalized} or post-training~\cite{lanchantin2025diverse, li2025preserving,li-etal-2025-jointly,yao2026diversity,chung2025modifying, chen2026posttraining} methods. Our work rather improves the diversity of already post-trained models. Specifically, we focus on short story generation and adopt a broad view of diversity that considers higher level narrative decisions. Figure~\ref{fig:story-diversity-collapse} illustrates why this broader view is necessary. Generations may differ in surface details, but maintain the same underlying narrative trajectory and other high level attributes. We define diverse short story generation as the task of producing a set of high quality stories for a single prompt such that they differ along salient narrative dimensions. This moves past conventional single-output generation, where the objective is to produce one fluent and instruction-following response. To reinforce this set property, we propose an optimization process that is based on specific group level rewards computed over multiple generations for the same prompt. This diversification methodology is captured by a two-phase training framework; we refer to it as \textbf{Div}ersified \textbf{L}anguage \textbf{M}odel or \divLM. First, we perform continued pre-training on creative writing data to strengthen the model's domain-specific generative capabilities. Then, we apply reinforcement learning with carefully curated group level rewards that encourage diversity while preserving basic output quality. Our approach promotes diversity over story level attributes (genre, style, tone, named entities) and maintains typical quality standards. Empirically, we find that optimizing story level diversity improves not only the targeted attributes, but also embedding based semantic distance. Our main contributions are summarized as follows:
\begin{enumerate}[leftmargin=*, topsep=-2.9pt, itemsep=-2.5pt]
\item We propose a two-phase training framework (\divLM) that combines continued pre-training on creative writing data with reinforcement learning based on group level diversity rewards.
\item We introduce a composite reward formulation that promotes diversity over story level attributes, including genre, style, tone, and named entities, while maintaining basic output quality.
\item Our analysis introduces novel metrics for quantifying diversity in LLM outputs. These go beyond cosine distance and are better tailored for the specificity of the short story generation task.
\item Our extensive experiments showcase that \divLM~yields significant gains on diversity performance while maintaining response quality.
\end{enumerate}

\section{Related Work}
\label{sec:related-work}
Prior work on LLM diversity can be broadly divided into methods applied during either model training or inference. The latter encourage diverse generations without updating model parameters, either by changing the decoding procedure or by modifying the prompt. Decoding based approaches modify the token selection process during generation, aiming to avoid overly deterministic or repetitive outputs while preserving relevance and coherence~\cite{franceschelli2025diffsampling,troshin-etal-2025-control}. Prompt based approaches instead induce diversity by changing the input context or task framing~\cite{wang-etal-2025-multilingual,chu-etal-2025-exploring,zhang2025verbalized}. While these methods intervene at inference time to induce more varied outputs, they do not directly modify the model's learned generation behavior. Training-time methods, on the other hand, incorporate diversity as an auxiliary objective during post-training procedures such as fine-tuning, reinforcement learning, and preference optimization. These approaches differ in how they define the diversity signal. One line of work uses lexical or token level proxies, encouraging variation in surface form or in the model's token distribution~\cite{li2025preserving,lanchantin2025diverse,yao2026diversity}. A second line of work measures diversity in a semantic representation space, encouraging generated responses to be farther apart according to embedding based criteria~\cite{li-etal-2025-jointly,chung2025modifying,chen2026posttraining}. These approaches move beyond lexical variation, but they typically compress diversity into a single semantic distance or equivalence signal. The most closely aligned work to ours uses reinforcement learning to encourage more diverse selections~\cite{anschel-etal-2025-group}. Here, diversity is defined over the distribution of selected options rather than over token choices or embedding distances between free form generations.

Despite this progress, existing diversity objectives remain limited for creative writing. Lexical and token level metrics primarily capture surface variation, while embedding based distances often collapse distinct creative dimensions into a single semantic similarity score. As a result, these approaches do not fully capture aspects of diversity that are central to creative generation. Moreover, much of the training time literature focuses on preserving the base model's diversity during post-training, whereas our work treats diversity as the primary target for improvement.

\section{Problem Formulation}
\label{subsec:problem-formulation}
Let $\mathcal{P}=\{p_1,\dots,p_N\}$ denote a set of $N$ story prompts. For each prompt $p_j$, a policy model $\pi_\theta$ generates a group of $G \ge 1$ outputs, $\mathcal{O}=\{o_1,\dots,o_G\}$. Each output $o_i$ is a short story conditioned on $p_j$ and obtains a reward score $\mathrm{R}\!\left(o_i\right)\in[-1,2]$, the derivation of which is detailed in section~\ref{subsec:reinforcement}. The objective is to optimize $\pi_\theta$ such that the generated group of outputs exhibits high diversity while maintaining acceptable quality. To this end, we treat diversity as the primary objective and enforce quality as a minimum requirement through the reward design. A condensed version of the optimization objective is given by
\begin{equation}
\label{eq:objective}
\argmax_{\theta} \, \mathbb{E}_{p_j \sim \mathcal{P},\mathcal{O}\sim\pi_{\theta'}(\mathcal{O}\mid p_j)}\left[ \frac{1}{G} \sum_{i=1}^{G} \Lambda_i\right] \, ,
\end{equation}
where $\theta'$ denotes the previous policy parameters, and $\Lambda_i \sim \mathrm{R}\left(o_i\right)$.

\section{Improving Diversity in Creative Writing with \divLM}
\label{sec:methodoloy}
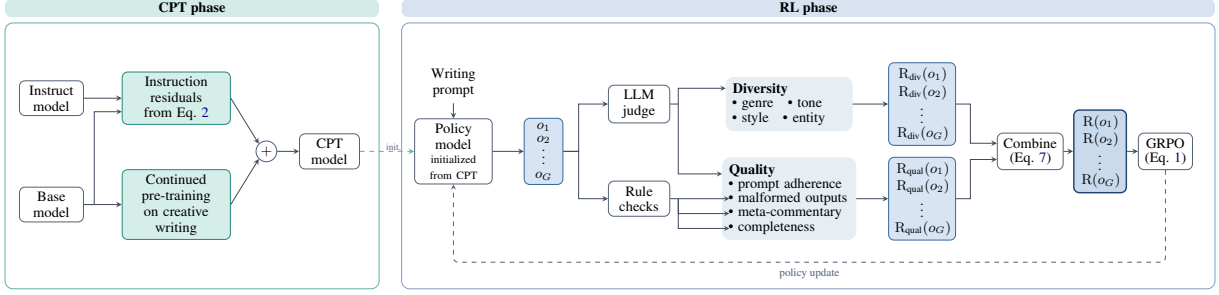
\begin{figure*}[t]
\centering
\input{figures/pipeline_tikz}
\caption{Overview of \divLM, our proposed framework for improving diversity in short story generation, consisting of the Continued Pre-Training (CPT, section~\ref{subsec:cpt}) and Reinforcement Learning (RL, section~\ref{subsec:reinforcement}) phases.}
\label{fig:pipeline}
\end{figure*}

The proposed methodology (\divLM) consists of two phases (Figure~\ref{fig:pipeline}). First, we perform continued pre-training to strengthen the model's foundational capabilities in creative writing. We then apply reinforcement learning to encourage greater output diversity while maintaining core requirements such as prompt adherence and completeness. 

\subsection{Continued Pre-training For Creative Writing (CPT Phase)} 
\label{subsec:cpt}
The objective of this phase is to expose the model to diverse creative writing examples so that it can more effectively generate varied and imaginative outputs in subsequent tasks. This will also enhance the effectiveness of reinforcement learning (section~\ref{subsec:reinforcement}) which depends on the capacity of the base model~\citep{guo2025deepseek} and can attain an enhanced performance ceiling via domain-specific continued pre-training~\citep{liu2025understanding}. Hence, we devise a domain-specific corpus that consists of creative writing texts curated to encourage stylistic flexibility and narrative variety (section~\ref{subsec:datasets}).

In our experiments, we perform continued pre-training using a low-rank adapter (LoRA), keeping the underlying model weights fixed~\citep{hu2022lora}. In line with standard practice, this is applied to the pre-trained base model~\citep{wang2024intruct, lu2025fine}. This process does not retain the instruction-following behavior of the instruction-tuned model~\citep{zhang2026instruction}, and hence these capabilities must be subsequently restored. To achieve this, we first compute an instruction residual $\Delta\theta$ by subtracting the parameters of the base model ($\theta_{\text{base}}$) from the instruction-tuned ones ($\theta_{\text{inst}}$)~\cite{jindal2024balancing}:
\begin{equation}
\label{eq:delta}
\Delta\theta = \theta_{\text{inst}} - \theta_{\text{base}} \, .
\end{equation}
Continued pre-training is then applied to the base model parameters, $\theta_{\text{base}}$, yielding $\hat{\theta}_{\text{base}}$. Finally, $\Delta\theta$ is added back to the updated parameters to recover the instruction-following behavior. According to our empirical analysis, an important design choice is to determine the proportion of $\Delta\theta$ that should be applied in order to recover instruction-following capabilities while preserving the diversity gains achieved during continued pre-training. We introduce a tunable scaling parameter $s\!\in\![0, 1]$ to control this. Therefore, the weights of the CPT model are given by
\begin{equation}
\label{eq:theta}
\hat{\theta}_{\text{base}}' = \hat{\theta}_{\text{base}} + s\Delta\theta \, .
\end{equation}

\subsection{Enhancing Diversity with Reinforcement Learning (RL phase)}
\label{subsec:reinforcement}
We employ Group Relative Policy Optimization (GRPO;~\citealp{shao2024deepseekmath}) on $\hat{\theta}_{\text{base}}'$ (Eq.~\ref{eq:theta}) to enhance output diversity. GRPO's objective aligns with the fact that diversity in LLM outputs can only be assessed by considering multiple responses per prompt. Furthermore, it reduces the critic-related memory and optimization overhead associated with Proximal Policy Optimization (PPO) style methods, as it foregoes a separately trained value model and directly exploits group level reward signals~\cite{shao2024deepseekmath}. For each prompt $p_j$, GRPO samples a group of outputs $\mathcal{O}=\{o_1,\dots,o_G\}$ from the previous policy $\pi_{\theta'}$. The exact optimization task requires setting
\begin{equation}
\begin{split}
    \label{eq:grpo}
    \Lambda_i = \min\!\left(\rho_i \, A_i,\, \left[\rho_i\right]_{\phi}^{\psi} A_i\right) - \beta \, \mathbb{D}_{\mathrm{\text{\scriptsize KL}}}\!\left(\pi_\theta \,\|\, \pi_{\mathrm{rf}}\right) \, ,
\end{split}
\end{equation}
in Eq.~\ref{eq:objective}, where $\pi_\theta$ and $\pi_{\mathrm{rf}}$ denote the current and reference (initialized from our CPT model) policy, $\rho_i = \pi_\theta\!\left(o_i \mid p_j\right) / \pi_{\theta'}\!\left(o_i \mid p_j\right)$, and $\left[\rho_i\right]_{\phi}^{\psi}$ denotes clipping from $\phi\!=\!1\!-\!\epsilon$ to $\psi\!=\!1\!+\!\epsilon$, where $\epsilon \geq 0$ is a hyperparameter. The advantage $A_i$ is the standardized reward $\mathrm{R}\!\left(o_i\right)$ calculated based on the reward values of all the outputs in the group. Finally, $\mathbb{D}_{\mathrm{\text{\scriptsize KL}}}\!\left(\pi_\theta \,\|\, \pi_{\mathrm{rf}}\right)$ penalizes divergence of the current from the reference policy, with the hyperparameter $\beta \geq 0$ controlling the strength of this regularization. Our reward formulation incorporates two components. The first focuses on \textbf{quality} requirements such as prompt adherence and completeness, while the other promotes output \textbf{diversity}.

\smallskip

\noindent\textbf{Quality Component.} We combine both LLM-based evaluations and heuristic checks to provide a comprehensive quality assessment. We consider prompt adherence ($\mathrm{R}_{\text{adhere}}$), absence of meta-commentary ($\mathrm{R}_{\text{meta}}$), and narrative completeness ($\mathrm{R}_{\text{comp}}$) as binary indicators, assigning scores that are either $1$ (true) or $0$ (false). Further details are provided in Appendix~\ref{sec:quality-metrics}. The initial reward score, $\mathrm{R}_{\text{qual}}$, for each generated output, $o_i$, is given by the average of the aforementioned three quality subscores. Outputs which are critically flawed, however, are assigned a fixed penalty that sets $\mathrm{R}_{\text{qual}}(o_i)=-1$. The determination of this is based on structural and linguistic quality criteria, such as length and coherence (Appendix~\ref{sec:quality-metrics}). This serves as a hard constraint to prevent the policy from reward hacking through trivially short or malformed text. Thus, the quality reward $\mathrm{R}_{\text{qual}}(o_i)\in\{-1, 0, \frac{1}{3}, \frac{2}{3}, 1\}$.

\smallskip

\noindent\textbf{Diversity Component.} The second component of the proposed reward function encourages diversity in generated short stories. While reference based metrics to measure lexical and semantic diversity can be used, they are vulnerable to reward hacking~\citep{skalse2022defining, pan2021the}. Moreover, prior work has shown that they often fail to capture nuanced, open-ended qualities such as style and narrative variation~\citep{li-etal-2025-generation}. We therefore only use them for evaluation and not as reward signals in GRPO training. 

Our reward function adopts a more abstract representation of narrative diversity that focuses of four dimensions: genre, style, tone, and named entities (including main characters). We deploy an LLM-as-a-judge to determine those.

For each one of genre, style, and tone diversity, we obtain a reward score as follows. Suppose we consider $K$ subtypes or categories (e.g., different genres). In order to achieve a uniform distribution of these subtypes that maximizes diversity across the $G$ prompt outputs/stories, each one must appear $G/K$ times, assuming that $\gamma\!=\!G\!\bmod\!K\!=\!0$. If $\gamma > 0$, the closest distribution to a uniform one will have $c_{\text{\tiny L}}\!=\!\lfloor G/K \rfloor$ instances of $K-\gamma$ subtypes and $c_{\text{\tiny H}}\!=\!\lceil G/K \rceil$ instances of $\gamma$ subtypes. For an output $o_i$, we count the amount of times, $\xi \leq G$, its subtype appears in the $G$ outputs. We then compute a proximity score, $\mathrm{S}(o_i)\in[0,1]$, which is a normalized deviation of $\xi$ from a uniform subtype distribution. It is given by:
\begin{equation}
\label{eq:proximity-score}
    \mathrm{S}(o_i) = 1 - \min\left(|\xi-c_{\text{\tiny L}}|,|\xi-c_{\text{\tiny H}}|\right)/d_{\max} \, ,
\end{equation}
where $d_{\max} = G - c_{\text{\tiny H}}$ is the largest possible deviation. Let $\eta \le K$ denote the number of subtypes in the $G$ outputs that have a count equal to $c_{\text{\tiny H}}$, and $\zeta$ denote the number of subtypes in the $G$ outputs that have a count equal to $c_{\text{\tiny L}}$. In the ideal case, $\eta = \gamma$ and $\zeta = K-\gamma$. When $\gamma > 0$, for outputs that are allocated a genre with total count $c_{\text{\tiny H}}$, their scores are corrected by applying a penalty 
\begin{equation}
    \mathrm{P}(o_i)=\max(0,(\eta - \gamma) / K) \, , 
\end{equation}
where $\mathrm{P}(o_i)\in[0,(K-\gamma) / K]$. Similarly, for outputs that are allocated a genre with total count $c_{\text{\tiny L}}$, their scores are corrected by applying a penalty
\begin{equation}
\mathrm{P}(o_i)=\max(0,(\zeta - K + \gamma) / K) \, ,
\end{equation}
where $\mathrm{P}(o_i)\in[0,\gamma / K]$. Thus, the final reward is calculated as $\mathrm{S}(o_i) - \mathrm{P}(o_i)$. Since the penalty $\mathrm{P}(o_i)$ is only applied when $\mathrm{S}(o_i)=1$, the reward lies in $[0,1]$. Rewards are calculated separately for genre, style, and tone, denoted by $\mathrm{R}_{\text{genre}}$, $\mathrm{R}_{\text{style}}$, and $\mathrm{R}_{\text{tone}}$. 

Named entity diversity operates over an open-ended rather than a fixed set of possible categories, and therefore requires a different reward formulation. We encourage named entities in each output $o_i$ to differ as much as possible from those in other outputs within the same group ($G$ outputs in total). Suppose that $o_i$ has a set of named entities $\mathcal{Q}_i$. The named entity reward $\mathrm{R}_{\text{ne}}(o_i) \in[0,1]$ is defined by
\begin{equation}
\label{eq:entity-score}
\mathrm{R}_{\text{ne}}(o_i)  = \left(1 - \frac{1}{|\mathcal{Q}_i|}\sum_{\nu\in\mathcal{Q}_i}\frac{\mathrm{g}(\nu) - 1}{G - 1}\right)^{\!h} \, ,
\end{equation}
where $\mathrm{g}(\nu)$ counts the number of outputs ($\leq G$) that contain entity $\nu\!\in\!\mathcal{Q}_i$, and $h>0$ is a tunable parameter that controls the sensitivity of the reward. To prevent the model from exploiting $\mathrm{R}_{\text{ne}}$ by producing a disproportionally large amount of named entities, we threshold both the denominator and the summation term of Eq.~\ref{eq:entity-score} to a value $u\!\in\!\mathbb{N}$ which is equal to the expected amount of entities produced by the original instruct model ($\theta_{\text{inst}}$) on a held-out validation set. Thus, we replace $|\mathcal{Q}_i|$ with $\min(|\mathcal{Q}_i|,u)$ and $\sum_{\nu\in\mathcal{Q}_i}\frac{\mathrm{g}(\nu)-1}{G-1}$ with $\min\left(\sum_{\nu\in\mathcal{Q}_i}\frac{\mathrm{g}(\nu)-1}{G-1},u\right)$. This calibrates the clipping threshold to the baseline entity count, preventing outputs with unusually many named entities from weakening the normalization effect. For outputs that contain no named entities ($|\mathcal{Q}_i|\!=\!0$), we assign a neutral score, equal to the mean $\mathrm{R}_{\text{ne}}$ across the outputs that contain at least one. Thus, the model is not encouraged to omit or introduce named entities unnecessarily. When no named entities exist in the $G$ outputs, we assign $\mathrm{R}_{\text{ne}}(o_i)\!=\!0.5$ to all, so that the entity diversity reward introduces no relative preference among them.
 
Combining all the above steps, the total diversity reward for an output $o_i$, $\mathrm{R}_{\text{div}}(o_i)\in[0,1]$, is obtained by computing the average across $\mathrm{R}_{\text{genre}}(o_i)$, $\mathrm{R}_{\text{style}}(o_i)$, $\mathrm{R}_{\text{tone}}(o_i)$, and $\mathrm{R}_{\text{ne}}(o_i)$. 

\noindent\textbf{Total RL Reward.} The final reward combines the quality and diversity components as follows:
\begin{equation}
\label{eq:total_reward}
\mathrm{R}(o_i) =
\begin{cases}
\mathrm{R}_{\text{qual}}(o_i) & \!\! \mathrm{R}_{\text{qual}}(o_i)\!<\!\tau\\
\mathrm{R}_{\text{qual}}(o_i)\!+\!\mathrm{R}_{\text{div}}(o_i) & \!\! \text{otherwise} \, ,
\end{cases}
\end{equation}
where $\tau\!\leq\!1$ is a predefined quality threshold calibrated on a validation set to reflect the minimum quality acceptable for the task. As discussed previously, $\mathrm{R}_{\text{qual}}(o_i)\in[-1,1]$ and $\mathrm{R}_{\text{div}}(o_i)\in[0,1]$. Therefore, $\mathrm{R}(o_i)\in[-1, \tau)$ when $\mathrm{R}_{\text{qual}}(o_i) < \tau$, and $\mathrm{R}(o_i)\in[\tau, 2]$ otherwise. Thus, the total reward $\mathrm{R}(o_i)\in[-1, 2]$. 

To prevent RL training from encouraging shorter outputs, we incorporate a length penalty into the total reward. This maintains output lengths comparable to those of the instruct model (see Appendix~\ref{sec:length-penalty}). The length penalty takes values in $[0,1]$ and is subtracted from the total reward only for non malformed outputs (i.e., when $\mathrm{R}(o_i)\in[0,2]$). Thus, the total reward $\mathrm{R}(o_i)$ remains in the range $[-1,2]$ after incorporating the length penalty.

The formulation in Eq.~\ref{eq:total_reward} ensures that diversity is only rewarded once a minimum level of structural quality is achieved. As noted by \citet{yao2026diversity}, incorrect outputs may artificially increase their reward by exploiting the diversity term. By withholding the diversity reward when quality is below the threshold, we prevent favoring low quality but superficially diverse outputs. If $\mathrm{R}_{\text{qual}}(o_i) < \tau$, the model is trained based only on a quality driven signal. Importantly, this does not impose a hard penalty (e.g., zero reward), but instead provides a graded signal that enables incremental improvement in output quality. When the threshold is reached, the diversity reward $\mathrm{R}_{\text{div}}(o_i)$ is added to the total reward. This encourages the exploration of diverse outputs while still receiving quality signals. 

This modular design of the reward function naturally induces a form of curriculum learning~\citep{bengio2009curriculum}. Early in training, when outputs are more likely to fall below the threshold, the model optimizes a simpler objective focused on quality. As it learns to consistently exceed the threshold, the objective expands to include diversity, resulting in a richer optimization target. This staged objective decomposition is consistent with recent approaches in which prerequisite behaviors are learned before more complex objectives are optimized within the same training pipeline~\citep{gao2024honest, jiang-etal-2025-teach}. 

\section{Experimental Results}
We use two open-source instruction-tuned models, \texttt{Llama-3.1-8B}~\citep{grattafiori-etal-2024-llama3} and \texttt{Gemma-2-9B}~\citep{gemma-team-2024-gemma2}. We first describe the datasets and evaluation metrics, then analyze the effects of the two training phases (CPT and RL) as well as provide further insights via ablation.

\subsection{Datasets} 
\label{subsec:datasets}
For the CPT phase (Section~\ref{subsec:cpt}), we constructed an 18M token corpus consisting of long form creative writing from Project Gutenberg\footnote{Project Gutenberg, \href{https://www.gutenberg.org/}{\nolinkurl{gutenberg.org}}} and \texttt{BookCorpus}~\citep{Zhu_2015_ICCV}. Original corpora from both resources include full length novels or stories across multiple genres and historical periods. To transform this material into a high quality training corpus, we deployed a multi-stage curation pipeline consisting of language filtering, toxicity screening, and deduplication (see Appendix~\ref{subsec:data-cpt}). For the RL phase (Section~\ref{subsec:reinforcement}) as well as for the evaluation of all methods, we use the \texttt{WritingPrompts} dataset that contains human-written prompts and short stories~\citep{fan-etal-2018-hierarchical}. Original data consists of posts (used as prompts) and their associated comments, each one being a story. Each prompt or story has a score that reflects community reception, calculated as the sum of all upvotes and downvotes received from community members. We filtered the dataset based on this score. Additionally, we removed prompts with high semantic similarity, and applied balanced sampling across implied genres, styles, and tones (see also Appendix~\ref{subsec:data-rl-eval}).

\subsection{Diversity Categories and Judge Validation}
\label{subsec:categorie-judge}
We have derived 9 genre categories from an empirical analysis of the book cataloging website Goodreads.\footnote{Goodreads, \href{https://www.goodreads.com/}{\nolinkurl{goodreads.com}}} Style is represented using 3 modes: descriptive, action focused, and dialogue heavy, following prior narrative mode taxonomies and controllable generation settings~\citep{selgin2009cunning, morrell2006between, marshall1998marshall, zhong-etal-2023-fiction}; see also Appendix~\ref{sec:writing-styles}. Tone (humorous, somber, neutral) captures the overall emotional register of the writing (see Appendix~\ref{sec:genre-tone-style} for a more detailed overview).

To assess the reliability of the LLM judge, we compare its annotations with human ones on a subset of the data. The judge-human agreement rate is 89.86\% which is quite similar to the  human-human agreement of 91.63\% (see Appendix~\ref{sec:llm-as-a-judge}).

\subsection{Evaluation Metrics and Experimental Settings}
\label{subsec:eval-metrics}
We evaluate generated outputs using metrics that quantify quality and diversity. All metrics are in the range of $[0,1]$. Diversity metrics are computed at the output group level ($G$ outputs for each prompt). We report both the RL reward and the coverage of genre, tone, and style. Coverage is defined as the average proportion of subtypes that are present in each group of outputs. We also report the named entity reward score. To compare with a conventional baseline metric, we include the average pairwise cosine distance between the embeddings of the outputs in each group. To ensure consistency with the scales of the other metrics, we rescale cosine similarity from $[-1,1]$ to $[0,1]$ and derive cosine distance from the rescaled values, resulting in a cosine distance scale of $[0,1]$.
Quality metrics are computed at the individual output level. They include prompt adherence, meta-commentary rate, completeness, and coherence (see Appendix~\ref{sec:quality-metrics}).

We create validation and test sets of 100 prompts each from the curated \texttt{WritingPrompts} dataset. We select prompts that maximize semantic diversity and maintain balanced representation across implied genres, styles, and tones (see Appendix~\ref{sec:dataset-preprocessing}). We use the validation set to select the hyperparameters and configuration settings for our experiments. CPT is performed on the corpus described in Section~\ref{subsec:datasets}, while reinforcement learning is performed using 3{,}000 training prompts with a group size of $G=10$. We tune the scaling parameter $s\in[0,1]$ on the validation set by evaluating values in increments of $0.1$. The named entity sensitivity parameter $h$ is evaluated over $\{1,2,3,4,5\}$. The quality threshold $\tau$ can take values in $\{-1,0,1/3,2/3,1\}$, of which we evaluate $\tau\in\{2/3,1\}$, excluding $\tau\leq1/3$ as overly lenient because it permits two of the three quality metrics to fail. Lastly, in GRPO (Eq.~\ref{eq:grpo}) the KL regularization coefficient $\beta$ is evaluated over $\{0.1,0.01,0.001\}$ and the clipping hyperparameter $\epsilon$ is set to $0.2$, consistent with established practice in RL literature~\citep{schulman2017proximal, yu2025dapo}.\footnote{TRL, \href{https://github.com/huggingface/trl}{\nolinkurl{github.com/huggingface/trl}}} Based on the combined quality and diversity reward on the validation set, we select $s=0.6$, $\tau=2/3$, $h=5$, and $\beta=0.001$ for the final experiments. We evaluate each model on the 100 test prompts, generating $G=10$ outputs per prompt, for a total of 1{,}000 outputs per model. Additional configuration details are provided in Appendix~\ref{sec:train-config}.

\subsection{Model Performance}
\label{subsec:perf}
\input{tables/main-perf.tex}

Table~\ref{tab:final-results} enumerates the test performance metrics for the baseline instruct models (INS), compared to results obtained after CPT, and the final outcomes that encompass both phases of our proposed \divLM. We also provide comparisons with other methods from the literature, namely Diversity Quality Optimization (DQO; \citealp{chen2026posttraining}) and Diversified Direct Preference Optimization (DDPO; \citealp{chung2025modifying}) using our datasets and  \texttt{Llama-3.1-8B} (see Appendix~\ref{sec:train-config}). 

CPT models outperform INS models across all diversity metrics. Prompt adherence and coherence remain comparable, while completeness decreases substantially, resulting in a lower average quality score. These results suggest that while CPT broadens the model's output distribution, the subsequent reinforcement learning phase is necessary to recover output quality.

\divLM~models yield quality metrics that remain close to those of the INS models, while completeness improves substantially. All diversity metrics improve. RL rewards respectively increase on average by 86.5\% and 42\% for Llama and Gemma compared to their INS baselines; respective gains over CPT are 23.7\% and 9.7\%. Smaller gains for Gemma may be partly due the fact that the hyperparameters were only optimized for  Llama.\footnote{To accommodate experiments under tight computational cost (GPU hours) constraints.} Therefore, the Gemma experiment acts as a transfer sanity check for \divLM, not a full cross-family baseline study. Nevertheless, the consistent improvement observed for Gemma suggests that the approach transfers across model families. A similar pattern of gains is observed in the coverage metrics, where CPT improves INS by 23.23\% and \divLM~improves over CPT by 20.65\%. The cosine distance also increases, which indicates greater semantic variation despite not being directly optimized during training. The human-written stories associated with the test prompts have an average cosine distance of .297. Relative to this human reference, the Llama INS model achieves 42.1\% of the observed semantic variation, while Llama \divLM~reaches 76.2\%; for Gemma, this increases from 41.8\% to 60.4\%. Hence, \divLM~substantially narrows the gap between human and LLM-generated outputs. 

From the baselines, the DQO model achieves high output quality, with an average score of .983. Compared to the INS model, it exhibits greater genre diversity and cosine distance but lower tone and named entity diversity. In contrast, the DDPO model improves across all diversity metrics relative to the INS model and achieves the highest cosine distance among all evaluated models. However, these gains come at a substantial cost to output quality, with DDPO obtaining the lowest average quality score of .534 and a completeness rate of only .017. Compared to DQO, \divLM~improves quality by 0.71\%, genre/tone/style coverage by an average of 50.34\%, and diversity rewards by 85.53\%. The respective gains compared to DDPO are 85.39\%, 8.91\%, and 10.50\%. Table~\ref{tab:five-run-results} enumerates performance metrics for a multi-run evaluation, highlighting the robustness of the obtained results.

\subsection{Human Evaluation}
\label{subsec:human-eval}
To independently validate the results obtained using our proposed metrics, we conduct a human evaluation of \divLM, DDPO, and DQO. The evaluation was conducted on 30 prompts with $G=5$, resulting in 150 stories per model. For each prompt-model pair, 3 human evaluators assessed the set of 5 generated stories, assigning a quality score and a diversity score, each on a 1-5 scale. We then averaged these scores across evaluators and prompts for each model (Table~\ref{tab:human-eval-results}). \divLM~achieves the best quality and diversity scores. In terms of quality, DQO performs similarly to \divLM, while DDPO scores substantially lower. For diversity, DDPO ranks second, while DQO ranks last. These results are consistent with those obtained using our proposed quality and diversity metrics.

\begin{table}[t]
\centering
\caption{Human evaluation results for \divLM, DDPO, and DQO in terms of quality and diversity, scored on a 1-5 scale.
Greater ($\uparrow$) values indicate better performance.} 
\label{tab:human-eval-results}
\setlength{\tabcolsep}{6pt}
\small
\begin{tabular}{lccc}
\toprule
& \multicolumn{3}{c}{\texttt{Llama-3.1-8B}} \\
\cmidrule(lr){2-4}
\bf Metric
& \multicolumn{1}{c}{DQO}
& \multicolumn{1}{c}{DDPO}
& \multicolumn{1}{c}{\divLM} \\
\midrule
Quality ($\uparrow$)
    & 4.006 & 1.794 & \bf 4.017 \\
Diversity ($\uparrow$)
    & 2.823 & 2.950 & \bf 3.872 \\
\midrule
\sc{Avg} ($\uparrow$)
    & 3.414 & 2.372 & \bf 3.944 \\
\bottomrule
\end{tabular}
\end{table} 

Notably, DDPO receives a lower relative diversity score in the human evaluation than in the automated evaluation. In the former, \divLM~improves diversity by 31.25\% compared to DDPO, whereas in the latter, this is reduced to 10.50\%. One possible explanation for this discrepancy is that human judgments of diversity are influenced by output quality. Human evaluators may be less likely to assign a high diversity score to a set of outputs when the individual stories are of poor quality, even if the outputs are different from one another. Therefore, aligning automated diversity evaluation more closely with human judgments may require conditioning diversity on quality. Despite this difference, the diversity reward aligns more closely with human judgments than cosine distance does, particularly in the ranking of the methods. This suggests that separation in a pre-trained embedding space may not always reflect meaningful narrative variation. We therefore include cosine distance as a complementary, judge-independent metric rather than treating it as a definitive measure of diversity.

\subsection{Instruction Residual Scaling Analysis}
\label{subsec:delta-theta-analysis}
\begin{figure}[b!]
\centering
\includegraphics[width=1.0\linewidth]{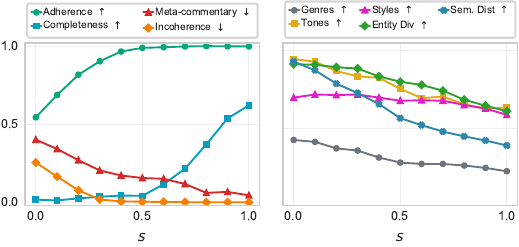}
\caption{Quality (left) and diversity (right) evaluation results for increasing proportions of the instruction residual ($\Delta\theta$) using \texttt{Llama-3.1-8B}.}
\label{fig:delta_scale}
\end{figure}

Using \texttt{Llama-3.1-8B}, we assess the performance of CPT models that use an increasing amount of the instruction residual ($\Delta\theta$), controlled by $s\!\in\![0, 1]$ in increments of $.1$. The results (Figures~\ref{fig:delta_scale},~\ref{fig:delta_scale_valid}, and~\ref{fig:delta_scale_genre_tone_style}) show a clear trade-off between output quality and diversity. As $s$ increases, adherence and completeness improve substantially, while meta-commentary and incoherence decrease. Adding more of the instruction residual makes the model more reliable, but also shifts it back toward the less diverse behavior of the INS model.

Continuing this experiment into the RL phase, we observe that lower values of $s$ result in lower initial quality scores, making it more difficult for the model to reach and maintain the quality threshold (Figure~\ref{fig:delta_scale_curve}). For $s\!\in\![0.0, 0.4]$, the model either never reaches the threshold or fails to remain above it and therefore does not sustain training for diversity. For $s\!\in\![0.5, 0.9]$, the model eventually crosses and maintains the quality threshold, allowing it to train successfully for diversity. At $s\!=\!1$, the model begins with the highest quality score; however, it is significantly more constrained, leaving insufficient flexibility for diversity optimization. Therefore, our empirical analysis indicates that a reasonable validation space for $s$ is in the $[0.5, 0.9]$ range.

\subsection{Ablation Study}
\label{subsec:ablations}

\begin{table}[t]
\centering
\caption{Ablation study based on \texttt{Llama-3.1-8B}. We compare the following model variants: Instruct (INS), GRPO-only, \grpocos, and \divLM~(shortened to \divLMshort) across quality and diversity metrics ranging in $[0,1]$. Greater ($\uparrow$) values indicate better performance.}
\label{tab:ablation-results}
\setlength{\tabcolsep}{2pt}
\small
\begin{tabular}{clllll}
\toprule
& & \multicolumn{4}{c}{\texttt{Llama-3.1-8B}} \\
\cmidrule(lr){3-6}
& \bf Metric
& \multicolumn{1}{c}{INS}
& \multicolumn{1}{c}{GRPO}
& \multicolumn{1}{c}{\grpocos}
& \multicolumn{1}{c}{\divLMshort} \\
\midrule
\multirow{5}{*}[-1ex]{\rotatebox{90}{\bf Quality}} &
Coherence
    & .997 & .999 & .999 & .997 \\
& Prompt adherence
    & .997 & .996 & .987 & .981 \\
& No meta-commentary
    & .968 & .996 & .999 & .998 \\
& Completeness
    & .726 & .986 & .850 & .985 \\
\cmidrule(lr){3-6}
& \sc{Avg}
    & .922 & \bf .994 & .959 & .990 \\
\midrule
\multirow{10}{*}[-2ex]{\rotatebox{90}{\bf Diversity}} &
Genre coverage
    & .210 & .250 & .319 & .382 \\
& Tone coverage
    & .603 & .603 & .616 & .832 \\
& Style coverage
    & .542 & .582 & .539 & .801 \\
\cmidrule(lr){3-6}
& \sc{Avg}
    & .452 & .478 & .491 & \bf .672 \\
\cmidrule(lr){2-6}
& $\mu\left(\mathrm{R}_{\text{genre}}\right)$
    & .293 & .383 & .546 & .606 \\
& $\mu\left(\mathrm{R}_{\text{tone}}\right)$
    & .345 & .345 & .360 & .667 \\
& $\mu\left(\mathrm{R}_{\text{style}}\right)$
    & .291 & .349 & .237 & .644 \\
& $\mu\left(\mathrm{R}_{\text{ne}}\right)$
    & .581 & .712 & .800 & .902 \\
\cmidrule(lr){3-6}
& \sc{Avg}
    & .378 & .447 & .486 & \bf .705 \\
\cmidrule(lr){2-6}
& Cosine distance
    & .125 & .167 & \bf .230 & .226 \\
\bottomrule
\end{tabular}
\end{table}

To assess the contribution of each component of our methodology, we evaluate two ablated variants. First, we apply GRPO training (RL phase) directly to the INS model without conducting CPT (GPRO-only). We also train an alternative \divLM~variant in which our custom diversity reward is replaced with pairwise cosine distance (\grpocos). The results are enumerated in Table~\ref{tab:ablation-results}.

The GRPO-only model marginally achieves the highest average quality score. This can be explained by the fact that training begins directly from the INS model, which already has strong quality performance, and the GRPO-only model  primarily needs to improve completeness and reduce meta-commentary. However, it also yields the smallest diversity gain among the trained variants, increasing the average diversity reward by only 18.3\% relative to the INS baseline. This indicates that GRPO can improve diversity without CPT, but its effect is limited when optimization begins from the more constrained INS model. Our interpretation is that CPT expands the set of responses available to the model, giving GRPO a broader space of trajectories for optimization.

The \grpocos~baseline improves average diversity rewards by 28.6\% compared to the INS model and achieves the highest semantic distance among all variants, as expected from its training objective. However, its average diversity reward remains below that of its CPT initialization. In particular, its tone and style rewards decrease substantially, suggesting that optimizing semantic distance alone does not reliably improve the broader range of diversity dimensions captured by our reward. Notably, the semantic distance achieved by \grpocos~is only 1.7\% higher than that of \divLM. This indicates that, despite not directly optimizing cosine distance, \divLM\ achieves comparable semantic variation while substantially improving diversity across all target dimensions.

We conduct complementary ablation analyses investigating alternative formulations of the diversity rewards in Appendices~\ref{sec:diversity-reward}~and~\ref{sec:entity-reward}. The results demonstrate that our proposed formulations outperform the alternatives considered. Furthermore, we investigate the effects of group size and decoding temperature in Appendices~\ref{subsec:group-size}~and~\ref{subsec:decoding-temp}, showing that our diversity improvements remain consistent across different group sizes and decoding temperatures. Finally, we analyze the sensitivity parameter $h$ in the named entity diversity reward in Appendix~\ref{subsec:h_ne}, finding that larger values generally lead to greater improvements in named entity diversity.

\section{Conclusion}
In this paper, we address the limited diversity of LLM outputs in short story generation, with a focus on four narrative dimensions, namely genre, tone, style, and named entities. We introduce \divLM, a two-phase post-training framework that combines CPT with diversity focused RL. In the CPT phase, we train the model on a creative writing corpus and use weight residuals to transfer instruction-following capabilities. We observe that the proportion of the residuals added controls the trade-off between output diversity and quality. During RL, we optimize the model using a custom composite reward that encourages variation across the four narrative dimensions while maintaining overall response quality. Our results show that \divLM~improves diversity significantly both over the instruct model variant as well as competitive alternative approaches. Further analysis highlights the complementary roles of the two training phases, wherein CPT broadens the model's output distribution, providing RL with a richer trajectory space to explore and optimize.

\section*{Limitations}
This study focuses exclusively on short story generation and does not experiment with other forms of creative writing, such as long-form stories and extended role-playing sessions. These tasks introduce additional requirements such as narrative continuity and long-term memory, which may interact with diversity in ways that are not captured by our current evaluation framework. Future work could investigate whether the proposed training approach remains effective in longer and more interactive generation settings, and extend the reward design to consider these additional dimensions.

Due to computational and memory constraints, our experiments are limited to models with at most 9B parameters. Evaluating \divLM~on larger models would provide a more complete understanding of its effectiveness and generalizability.

Finally, this work addresses output diversity rather than creativity in its broader sense. Diversity involves the degree of variation across generated outputs and represents only one component of creative behavior. Developing reliable methods to evaluate and optimize creativity comprehensively remains an important direction for future work.

\section*{Acknowledgments}
We would like to acknowledge the support from UKRI's AIRR Gateway Programme providing access to GPU computational support via the Isambard AI cluster (project titled ``Foundation models for sequential predictions''). Both authors would like to acknowledge the feedback provided by Jayroop Ramesh during various stages of this work. VL would like to acknowledge all levels of support from the UKRI-funded Science of Fundamental AI Research (SOFAIR) Lab.

\FloatBarrier

\nocite{selgin2009cunning,morrell2006between,marshall1998marshall,zhong-etal-2023-fiction}
\bibliography{custom}

\clearpage

\appendix
\setcounter{table}{0}
\renewcommand{\thetable}{S\arabic{table}}%
\setcounter{figure}{0}
\renewcommand{\thefigure}{S\arabic{figure}}%
\setcounter{equation}{0}
\renewcommand{\theequation}{S\arabic{equation}}%

\section*{\Large\centering Appendix}

\section{About Writing Styles}
\label{sec:writing-styles}
What we refer to as \emph{style} is often termed as \emph{narrative mode} in the creative writing and stylistics literature. Prior work has proposed several taxonomies of fiction writing modes, which overlap substantially. \citet{selgin2009cunning} identifies five modes: action, dialogue, thoughts, summary, and description. \citet{morrell2006between} expands to action, exposition, description, dialogue, summary, and transition. Similarly,~\citet{marshall1998marshall} lists action, summary, dialogue, feelings/thoughts, and background as core modes. These taxonomies consistently distinguish between physical action, spoken dialogue, and narratorial or descriptive exposition, while differing mainly in whether internal thought, exposition, summary, and background information are treated as separate categories. To this end, we have chosen to follow the paradigm of prior work in controllable fiction generation that employs dialogue, action, and description as explicit writing modes~\citep{zhong-etal-2023-fiction}.

\section{Construction of Genre, Style, and Tone Dimensions}
\label{sec:genre-tone-style}
We construct the diversity dimensions (genre, style, and tone) to be as mutually distinguishable and semantically orthogonal as possible, while remaining interpretable to both human readers and LLM-based judges. Each dimension consists of a small set of discrete categories designed to support robust LLM judgment and balanced reward allocation.

\smallskip

\noindent\textbf{Genre.} The genre categories were derived empirically from an analysis of 10{,}634 books and their associated genres as reported on the book cataloging website Goodreads. We first removed all non-fiction genres (e.g., philosophy, religion, biography, politics). We then computed pairwise genre overlap and merged categories with high overlap (Jaccard index $\geq 0.2$). Finally, we removed genres that primarily denote demographics or time periods (e.g., young adult, 20th century). The resulting set reflects broad narrative genres with minimal semantic redundancy: \emph{fantasy}, \emph{science fiction}, \emph{horror}, \emph{romance}, \emph{mystery/thriller}, \emph{literary drama}, \emph{historical}, \emph{action/adventure}, and \emph{other}. The \emph{other} category captures works that do not fit into the preceding genres or exhibit mixed or less common characteristics (e.g., dystopian fiction or magical realism).

\noindent\textbf{Style (Narrative Mode).} To reduce overlap and improve robustness of automatic classification by an LLM judge, we collapse the finer-grained distinctions (see Appendix~\ref{sec:writing-styles}) into three broad modes: \emph{descriptive}, which includes narration, exposition, description, and background;
\emph{action focused}, emphasizing physical events and movement;
and \emph{dialogue heavy}, in which story progression occurs primarily through character speech. This formulation is consistent with prior work in controllable fiction generation, which uses dialogue, action, and description as controllable writing modes~\citep{zhong-etal-2023-fiction}. Although individual stories may exhibit elements of multiple modes, each story is assigned a single label that identifies its dominant narrative style.

\noindent\textbf{Tone.} We define tone using three categories: \emph{humorous}, \emph{somber}, and \emph{neutral}. Rather than treating humor as a separate genre, we assume that tone is orthogonal to genre, allowing any genre to be expressed in a humorous, somber, or neutral manner.

\smallskip

The aforementioned dimensions define a compact and interpretable space over which diversity can be enforced, addressing aspects of creative writing that are not always captured by surface level similarity metrics.

\section{Corpora Preprocessing}
\label{sec:dataset-preprocessing}

\subsection{Creative Writing Dataset}
\label{subsec:data-cpt}
To form the dataset for continued pre-training, we first curate the selection of 150 books to ensure a uniform distribution across genres. Books are obtained from both Project Gutenberg and the \texttt{BookCorpus}~\citep{Zhu_2015_ICCV} dataset (MIT license). We filter the books to retain only the narrative text of the story. We remove front and back matter (e.g., tables of contents), boilerplate (e.g., ``\texttt{*** END OF THE PROJECT GUTENBERG EBOOK ***}''), illustration markers and captions (e.g., ``\texttt{[Illustration]}''), and formatting artifacts, such as decorative separators (e.g., ``\texttt{*****}'') and layout markers (e.g., ``\texttt{/* RIGHT ... */}''). We identify and remove these elements using rule-based pattern matching. The texts are segmented at sentence boundaries into passages of 500-800 tokens, with token counts measured using the \texttt{Llama-3.1-8B} tokenizer. Consecutive sentences are added until the next one would exceed the 800-token limit. We perform language filtering using the fastText language identification model (\citealp{joulin2016fasttext, joulin2016bag}; CC Attribution-Share-Alike 3.0), and retain only English texts. We apply automated toxicity screening to the BookCorpus dataset using the Detoxify library (Apache License 2.0)\footnote{Detoxify repository, \href{https://github.com/unitaryai/detoxify}{\nolinkurl{github.com/unitaryai/detoxify}}} to identify passages containing toxic or otherwise unsuitable content. We set the threshold values for explicit, obscene, and toxic content to 0.2, 0.6, and 0.6, respectively. Any passage exceeding these conservative thresholds in the corresponding toxicity dimensions is removed. This process filters out undesirable material without constraining the creative or stylistic diversity of the corpus. This filtering is not applied to books from Project Gutenberg. To remove redundant passages while preserving the natural variability of creative writing, we follow the approach of~\citet{lee-etal-2022-deduplicating} for deduplication. We use exact matching, substring overlap (via 50-character substrings), edit similarity ($> 0.7$) and Jaccard similarity ($> 0.7$) as primary signals, and add cosine similarity ($> 0.9$) as an additional safeguard for semantic duplicates. Although these thresholds are intentionally strict, the inherent diversity of creative prose means that only 0.6\% of passages are removed in this step.

\begin{figure}[t]
\centering
\small
\begin{tikzpicture}
\node[
    draw=gray!45,
    fill=gray!5,
    rounded corners=3pt,
    inner sep=7pt,
    text width=0.92\linewidth
] {
\textbf{Pair 1}

\vspace{2pt}
\textbf{(a)} ``9 out of 10 dentists recommend this toothpaste. You are the 1 out of 10 who doesn't -- and for a very good reason.''

\vspace{3pt}
\textbf{(b)} ``You disagree with most of your co-workers and peers. You hate toothpaste. You are the 10th Dentist.''

\vspace{7pt}
\textbf{Pair 2}

\vspace{2pt}
\textbf{(a)} ``You are an attorney. A client tells you he's found a genie and wants your help in crafting a wish with no ironic loopholes the genie can exploit. Your retainer fee is the third wish.''

\vspace{3pt}
\textbf{(b)} ``The genie has taken a liking to you. `Take this notebook. Make your wish as specific, loophole free as possible. I will return in one year.' You immediately hire a lawyer.''
};
\end{tikzpicture}
\caption{Examples of semantically similar prompts identified during deduplication.}
\label{fig:dedup-examples}
\end{figure}

\subsection{Writing Prompts Dataset} 
\label{subsec:data-rl-eval}
The writing prompts are obtained from the \texttt{WritingPrompts}~\citep{fan-etal-2018-hierarchical} dataset (MIT license). We first filter prompts based on the human-assigned popularity scores in the dataset. We assume that strong community engagement, in the form of upvotes and substantive story responses, serves as a proxy for prompt quality and narrative potential. Therefore, we retain prompts with a post score greater than 5, more than two comments, and at least one comment with a score greater than 2. To identify and remove similar prompts, we compute normalized sentence embeddings using \texttt{all-mpnet-base-v2}~\citep{song2020mpnet,reimers-gurevych-2019-sentence} and compute pairwise cosine similarities between all prompt embeddings. For prompt pairs with cosine similarity greater than 0.55, we remove one prompt from the pair. Examples of semantically similar prompts identified during this stage are shown in Figure~\ref{fig:dedup-examples}.

To prevent the dataset from being biased toward specific genres, tones, or styles, we design a balanced sampling pipeline. We prompt the judge LLM to classify prompts into one of three categories: constrained, implied, or open-ended. For prompts labeled as constrained or implied, the judge additionally assigns the corresponding genre, style, or tone labels. Finally, the judge is asked to assess prompt validity in order to remove prompts that cannot be meaningfully answered by a language model. Examples of invalid prompts include instructions that depend on external or unavailable context, such as ``Instead of this title, take any title from your Reddit frontpage as prompt'' or ``Write a story about this pic that made the front page of Reddit.'' The final dataset is then constructed by uniformly sampling valid prompts across the identified genres, styles, and tones to ensure balanced representation.

\begin{figure*}[t!]
\centering
\begin{tikzpicture}
  \node[anchor=south west, inner sep=0] (img) at (0,0)
    {\includegraphics[width=0.6\linewidth]{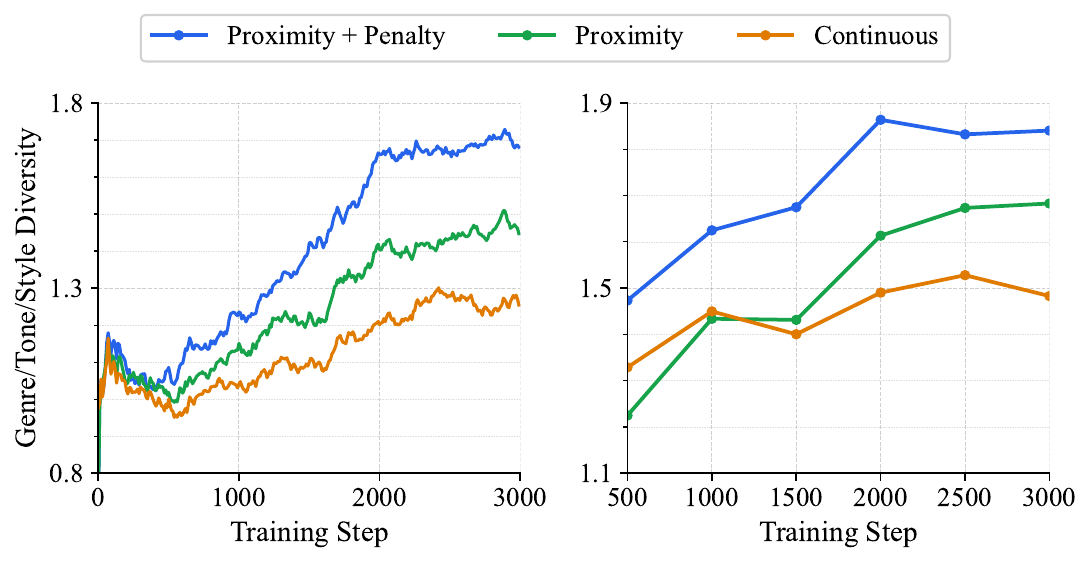}};
  \begin{scope}[x={(img.south east)}, y={(img.north west)}]

    \node at (0.30,-0.06) {\small (a)};
    \node at (0.80,-0.06) {\small (b)};
  \end{scope}
\end{tikzpicture}
\caption{Genre, tone, and style diversity curves for different reward formulations on Llama. The score is the sum of the genre, tone, and style rewards, resulting in a range of $[0,3]$. (a) Training scores; (b) Validation scores evaluated every 500 steps.}
\label{fig:diversity-reward-curves}
\end{figure*}

\section{Quality Metrics}
\label{sec:quality-metrics}
The quality metrics used during training are prompt adherence, meta-commentary, and completeness. We use \texttt{Gemma-3-12b-it} as a judge to evaluate these metrics~\citep{team2025gemma}. The judge is prompted with a structured system message defining evaluation criteria and a required JSON output format, ensuring consistent and parsable responses. Prompt adherence assesses whether the generated story satisfies the essential requirements specified in the prompt, including characters, setting, constraints, and stylistic elements (if explicitly stated). Meta-commentary captures the presence of undesirable meta-textual content, defined as any content that falls outside the narrative itself. This includes statements about the story (e.g., ``Here is your story'', ``In this story I will'') and comments directed at the user. Completeness evaluates whether the story forms a coherent and self-contained narrative. A complete story should have a clear beginning, development, and resolution. Outputs are penalized if they are truncated or end abruptly (e.g., mid-sentence or mid-scene). In addition to the LLM judge, we apply rule-based checks for meta-commentary and completeness. The meta-commentary check searches directly for common meta-textual phrases identified empirically, such as ``Is there anything else I can help you with?''. The completeness check uses a simple punctuation heuristic, flagging outputs that do not end with a punctuation mark as incomplete. For each of these metrics, an output is marked as failing if either the LLM judge or the corresponding rule-based check detects a violation. All three metrics are binary, i.e., a score of 1 indicates the desired behavior, while a score of 0 indicates failure. 

An output is classified as malformed if it violates basic structural or linguistic quality criteria. Specifically, we flag outputs that are insufficient in length (fewer than 50 words), exhibit low lexical diversity (a unique token ratio below 30\%), or contain formatting artifacts. Formatting artifacts include special token leakage (e.g., ``\texttt{\textless|im\_end|\textgreater}'', ``\texttt{\textless|endoftext|\textgreater}'') as well as HTML or CSS fragments indicative of formatting corruption (e.g., ``\texttt{font-size}'', ``\texttt{font-weight}''). Furthermore, we prompt the judge to assign binary scores for coherence and grammatical correctness. Outputs receiving a score of 0 on either metric are counted as malformed.

\section{Alternative Formulations for the Diversity Reward}
\label{sec:diversity-reward}
The diversity component of our methodology derives a diversity reward for genre, tone, and style (the reward for named entities is defined differently). We justify the choice of this reward, labeled as \textbf{Proximity + Penalty}, by comparing with two alternative, perhaps more straightforward formulations.

\smallskip

\noindent\textbf{Proximity.} This variant relies solely on the proximity score from Eq.~\ref{eq:proximity-score}:
\begin{equation}
\label{eq:reward-proximity}
\mathrm{R}(o_i) = \mathrm{S}(o_i) \, .
\end{equation}

\noindent\textbf{Continuous.} This variant defines the ideal count as $c = G/K$, regardless of whether $G \bmod K = 0$. Thus, $c$ may have a non integer value, and the reward is computed as
\begin{equation} 
\label{eq:continuous-score} 
\mathrm{R}(o_i) = 1 - |\xi-c|/d_{\max} \, . 
\end{equation}

\smallskip

We train the Llama model using each reward variant and report the training and validation curves (Figure~\ref{fig:diversity-reward-curves}) along with the final evaluation results (Table~\ref{tab:div-variants-result}). The `Proximity + Penalty' variant obtains the best genre, tone and style rewards.

\section{Named Entity Reward}
\label{sec:entity-reward}
We compare four formulations of the named entity reward. In all cases, $\mathcal{Q}_i$ denotes the set of normalized named entities extracted from output $o_i$, after removing any named entities that appear in the prompt. Named entities are extracted using our LLM judge model.

\begin{table}[b]
\centering
\caption{Reward performance for different diversity reward formulations on the test set. The score ($\in [0,3]$) is the sum of the genre, tone, and style rewards.}
\small
\setlength{\tabcolsep}{6pt}
\renewcommand{\arraystretch}{0.95}
\begin{tabular}{@{}lr@{}}
\toprule
Formulation & Score \\
\midrule
Continuous          & 1.634 \\
Proximity           & 1.712 \\
Proximity + Penalty & \textbf{1.917} \\
\bottomrule
\end{tabular}
\label{tab:div-variants-result}
\end{table}

\smallskip

\noindent\textbf{Original.} This variant uses the reward from Eq.~\ref{eq:entity-score} without clipping.

\noindent\textbf{Clipped.} This variant applies the clipping described in Section~\ref{subsec:reinforcement} to Eq.~\ref{eq:entity-score}. It is the reward used in our proposed model.

\noindent\textbf{Penalty.} This variant applies an entity count penalty to the original reward:
\begin{equation}
\label{eq:ne-penalty}
\mathrm{P}_{\text{ne}}(o_i) = \exp\left(-\left|\frac{|\mathcal{Q}_i|-\mu}{\sigma}\right|\right) \, ,
\end{equation}
where $\mu$ and $\sigma$ are the mean and standard deviation of $|\mathcal{Q}_i|$ estimated from the outputs of the INS model on the validation set. The penalty is applied by computing $\mathrm{R}_{\text{ne}}(o_i) \cdot \mathrm{P}_{\text{ne}}(o_i)$. This encourages the number of named entities to remain close to that of the baseline INS model, reducing the incentive to exploit the reward through unusually large or small entity counts.

\noindent\textbf{Exponential.} This variant removes the explicit normalization by $|\mathcal{Q}_i|$ and instead applies an exponential transformation directly to the summation term:
\begin{equation}
\label{eq:ne-exp}
\mathrm{R}_{\text{ne}}(o_i) = \exp\left(-\sum_{\nu\in\mathcal{Q}_i}\frac{\mathrm{g}(\nu)-1}{G-1}\right) \, .
\end{equation}

\smallskip

In all four variants, no entity outputs are assigned neutral scores as described in Section~\ref{subsec:reinforcement}. In the Penalty variant, $\mathrm{P}_{\text{ne}}(o_i)$ is computed and applied to no entity outputs after the neutral score is assigned. Table~\ref{tab:entity-diversity-exps} reports the average number of unique and repeated entities per group on the test set for each variant. The original variant incentivizes the model to increase the number of unique entities, producing more than twice as many as the baseline. The penalty variant mitigates this behavior but does not fully prevent it. Overall, the clipped variant is the only formulation that keeps the number of unique entities close to the baseline while significantly reducing the number of repeated entities.

\begin{table}[t]
\centering
\small
\setlength{\tabcolsep}{3pt}
\begin{tabular}{lrr}
\toprule
Variant & \makecell[r]{Avg. unique entities\\per group} & \makecell[r]{Avg. repeated entities\\per group} \\
\midrule
INS & 28 & 4.7 \\
Original & 62 & 2.8 \\
Exponential & 21 & 2.1 \\
\textbf{Clipped} & \textbf{26} & \textbf{1.5} \\
Penalty & 49 & 2.4 \\
\bottomrule
\end{tabular}
\caption{Results on the test set for different named entity diversity formulations with $G=10$ on the \texttt{Llama-3.1-8B} model. }
\label{tab:entity-diversity-exps}
\end{table}

\section{LLM-as-a-Judge Evaluation Based on Human Labels}
\label{sec:llm-as-a-judge}
The use of LLM-as-a-judge is an active area of methodological debate. Prior work has shown that LLM judges can approximate human preferences at scale, but can also be vulnerable to issues such as position and verbosity bias \citep{zheng2023llmjudge, gu-etal-2025-survey, li-etal-2025-generation}. To validate our evaluator, we manually annotate 100 samples, corresponding to 10\% of the 1000-sample test set. We annotate all samples for prompt adherence, completeness, meta-commentary, genre, style, and tone. We report both inter-annotator agreement and human-LLM agreement (Table~\ref{tab:human-judge-agreement}). Overall, human-human (91.6\%) and judge-human (89.9\%) agreement rates overlap significantly. Notably, inter-human agreement is substantially higher for the quality related than for the diversity related metrics. This is expected, as judgments of genre, tone, and style are more subjective and may allow for several reasonable interpretations. For the quality related metrics, judge-human agreement is slightly lower than inter-human agreement, indicating that the judge does not fully reproduce the relatively strong consensus among human annotators. In contrast, for the diversity related metrics, judge-human agreement is slightly higher than inter-human agreement. We note that judge-human agreement is measured against the aggregate human judgment rather than against each individual annotator. Hence, aggregation can reduce annotator specific variation. The judge may therefore agree more closely with the human consensus than individual humans agree with one another, particularly on the more subjective diversity related metrics.

\begin{table}[t]
\centering
\small
\setlength{\tabcolsep}{6pt}
\begin{tabular}{lrr}
\toprule
Metric & Human-human & Judge-human \\
\midrule
\multicolumn{3}{l}{\textbf{Quality metrics}} \\
Prompt adherence    & 92.30\%  & 90.40\% \\
Meta-commentary     & 97.60\%  & 95.70\% \\
Completeness        & 100.00\% & 94.10\% \\
Coherence           & 97.00\%  & 92.70\% \\
\sc{Avg}            & 96.73\%  & 93.23\% \\
\midrule
\multicolumn{3}{l}{\textbf{Diversity metrics}} \\
Genre               & 80.60\%   & 82.70\% \\
Tone                & 84.60\%   & 81.70\% \\
Style               & 89.30\%   & 91.70\% \\
\sc{Avg}            & 84.83\%  & 85.37\% \\
\midrule
Total \sc{Avg}      & 91.63\%  & 89.86\% \\
\bottomrule
\end{tabular}
\caption{Human-human and judge-human agreement rates for quality and diversity metrics.}
\label{tab:human-judge-agreement}
\end{table}

\section{Training Configuration}
\label{sec:train-config}
\noindent\textbf{\divLM.} In the continued pre-training phase, we conducted training using LoRA ($r=32$, $\alpha=32$) applied to attention and feed-forward modules. Training was conducted for one epoch with a batch size of 16. The learning rate was set to $1 \times 10^{-4}$  with cosine scheduling and 8\% warmup. After adding the instruction residual, we applied GRPO to the resulting model using LoRA ($r=16$, $\alpha=16$) on the attention and feed-forward modules. The learning rate was set to $4 \times 10^{-5}$ with cosine scheduling and 10\% warmup. We conducted training for two epochs on 3{,}000 prompts with a batch size of 20, and a group size $G=10$. Across all experiments, the maximum generation length was set to 700 tokens and the temperature was set to 1. The continued pre-training phase was conducted on a single NVIDIA GH200, while the GRPO training was performed on four GH200s. AdamW~\citep{loshchilov2018decoupled} optimizer was used in both training phases. The embedding model used to compute cosine distance was \texttt{all-mpnet-base-v2} (\citealp{song2020mpnet,reimers-gurevych-2019-sentence}; Apache License 2.0).

\begin{figure}[t!]
\centering
\includegraphics[width=1.0\linewidth]{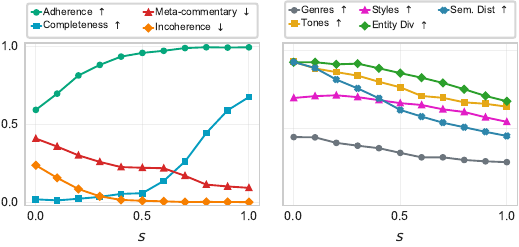}
\caption{Validation results for different instruction residual scales $s$ on \texttt{Llama-3.1-8B}. Diversity metrics consist of genre, tone, and style coverage values, as well as the named entity diversity reward $\mathrm{R}_{\text{ne}}$. All metrics are reported on a $[0,1]$ scale.}
\label{fig:delta_scale_valid}
\end{figure}

\noindent\textbf{DQO.} For DQO~\citep{chen2026posttraining}, we trained the \texttt{Llama-3.1-8B} INS model on our dataset of 3{,}000 samples using the authors' released DQO-PPO implementation.\footnote{DQO repository, \href{https://github.com/fairytale9/diversity-quality-optimization}{\nolinkurl{github.com/fairytale9/diversity-quality-optimization}}} We followed the authors' reported hyperparameters where applicable: three training epochs, an actor learning rate of $1 \times 10^{-6}$, a critic learning rate of $1 \times 10^{-5}$, a rollout temperature of 1, \texttt{Skywork-Reward-V2-Llama-3.2-1B} as the reward model, and \texttt{all-MiniLM-L6-v2} (Apache License 2.0) as the embedding model. To accommodate our larger model size, we reduced the batch size to 32. We set the maximum response length to 700 tokens for consistency with our other experiments. In addition to the PPO variant, we evaluated the GRPO variant of DQO. We selected the PPO variant for direct comparison with \divLM~as it achieved better overall results (Table~\ref{tab:dqo-grpo_ppo_comparison}).

\begin{table}[t!]
\centering
\caption{Performance comparison between the GRPO and PPO variants of DQO.}
\small
\setlength{\tabcolsep}{4pt}
\begin{tabular}{lrr}
\toprule
Metric & GRPO & PPO \\
\midrule
Average quality score
    & .894 & \textbf{.983} \\
Average Genre/Tone/Style coverage
    & \textbf{.454} & .447 \\
Average diversity reward
    & .360 & \textbf{.380} \\
Cosine distance
    & .118 & \textbf{.135} \\
\bottomrule
\end{tabular}
\label{tab:dqo-grpo_ppo_comparison}
\end{table}

\noindent\textbf{DDPO.} For DDPO~\citep{chung2025modifying}, we trained \texttt{Llama-3.1-8B} using the authors' released \texttt{DiversityTuning} implementation.\footnote{Diversity Tuning repository, \href{https://github.com/mj-storytelling/DiversityTuning}{\nolinkurl{github.com/mj-storytelling/DiversityTuning}}} Because the method requires human-written responses paired with the training prompts, we first retrieved the corresponding responses for our prompts. Following the authors' pipeline, we fine-tuned the model on the resulting dataset. We then initialized DDPO from this fine-tuned checkpoint and trained it on the authors' semantic and stylistic preference pairs constructed from the same dataset. Following the codebase, the DDPO stage used three training epochs, a learning rate of $5 \times 10^{-6}$, and $\beta = 0.1$ (not to be confused with GRPO's $\beta$ in Eq.~\ref{eq:grpo}). We set the batch size to 4 with a maximum response length of 700 tokens.

\section{Instruction Residual Scaling Analysis}
\label{sec:scaling-extra}
\begin{figure}[t!]
    \centering

    \begin{subfigure}[t]{0.49\linewidth}
        \centering
        \includegraphics[width=\linewidth]
        {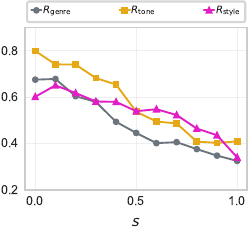}
        \caption{Test set}
        \label{fig:delta_scale_2}
    \end{subfigure}
    \hfill
    \begin{subfigure}[t]{0.49\linewidth}
        \centering
        \includegraphics[width=\linewidth]
        {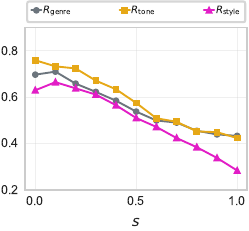}
        \caption{Validation set}
        \label{fig:delta_scale_2_valid}
    \end{subfigure}

    \caption{Genre, tone, and style rewards obtained for different instruction residual scales ($s$) on \texttt{Llama-3.1-8B}. Rewards are reported on a $[0,1]$ scale.}
    \label{fig:delta_scale_genre_tone_style}
\end{figure}

\begin{figure}[t!]
    \centering
    \includegraphics[width=0.8\linewidth]{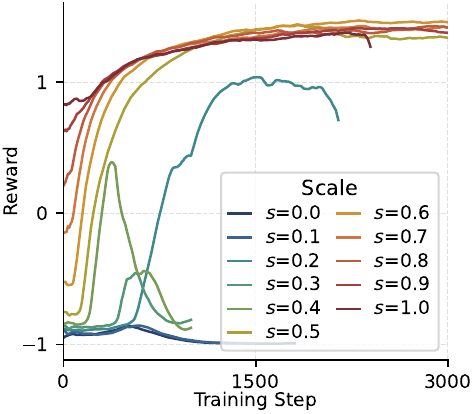}
    \caption{
        Training curves for \texttt{Llama-3.1-8B} under different residual scales ($s$), showing the total reward $R \in [-1,2]$ across training steps.
    }
    \label{fig:delta_scale_curve}
\end{figure}

In addition to the results on the test set for different values of $s$ (Figure~\ref{fig:delta_scale}), we report the corresponding results on the validation set in Figure~\ref{fig:delta_scale_valid} and the mean genre, tone, and style rewards on both the test and validation sets in Figure~\ref{fig:delta_scale_genre_tone_style}. The trends are consistent across all results, with quality improving and diversity decreasing as $s$ increases. Figure~\ref{fig:delta_scale_curve} additionally presents the RL training curves for each value of $s$.

\section{Sensitivity Parameter ($h$) in Named Entity Diversity Reward}
\label{subsec:h_ne}
\begin{figure}[b!]
\centering
\includegraphics[width=1.0\linewidth]{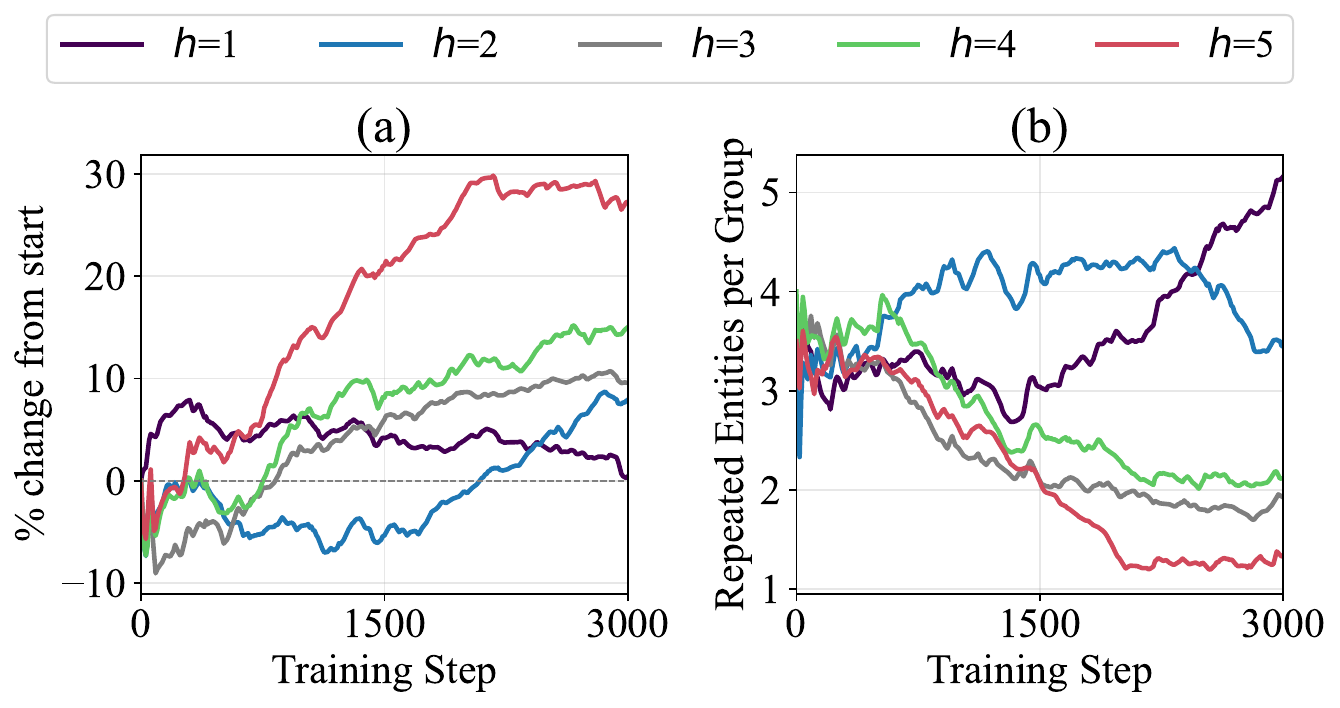}
\caption{Training curves for \texttt{Llama-3.1-8B} under different values of named entity reward sensitivity parameter ($h$). (a) Percentage change in the entity diversity reward over training. (b) Average number of repeated entities per group over training.}
\label{fig:entity_temp}
\end{figure}

We experiment with different values of the sensitivity parameter $h$ in the named entity diversity reward (Eq.~\ref{eq:entity-score}). To facilitate comparison across reward scales, we report the percentage change in reward relative to its initial value and the average number of repeated entities per group over training (Figure~\ref{fig:entity_temp}). At $h=1$, the reward provides insufficient pressure to reduce entity repetition. Although it increases slightly early in training, this improvement is not sustained, and the number of repeated entities ultimately increases. As $h$ increases, repeated entities are penalized more harshly, leaving greater scope for optimization. The best performance is achieved by setting $h=5$, which obtains the largest improvement over the initial reward (55.25\%) and the largest reduction in the number of repeated entities over total entities per group (50.47\%) as enumerated in Table~\ref{tab:entity_metrics_h1}. 

\begin{table}[t!]
\centering
\caption{Entity repetition ratio (repeated entities per group over total entities per group) and mean $\mathrm{R}_{\text{ne}}$ for INS and \divLM~across values of $h$. Final evaluation uses $h=5$ for consistency.}
\small
\setlength{\tabcolsep}{4pt}
\begin{tabular}{lrrrrrr}
\toprule
Metric & INS & $h=1$ & $h=2$ & $h=3$ & $h=4$ & $h=5$ \\
\midrule
Ratio $\downarrow$
    & .107 & .117 & .066 & .068 & .078 & \textbf{.053} \\
$\mu\!\left(\mathrm{R}_{\text{ne}}\right)$ $\uparrow$
    & .581 & .686 & .838 & .873 & .813 & \textbf{.902} \\
\bottomrule
\end{tabular}
\label{tab:entity_metrics_h1}
\end{table}

\begin{table}[t!]
\centering
\caption{Average cosine distance on the test set for INS and \divLM~ under different values of the decoding temperature ($T$).}
\small
\setlength{\tabcolsep}{3pt}
\renewcommand{\arraystretch}{0.95}
\begin{tabular}{@{}lrrrrr@{}}
\toprule
Model & $T=0.8$ & $T=0.9$ & $T=1.0$ & $T=1.1$ & $T=1.2$ \\
\midrule
INS
& .116 & .118 & .125 & .131 & .138 \\
\divLM
& \textbf{.215} & \textbf{.225} & \textbf{.226}
& \textbf{.231} & \textbf{.236} \\
\bottomrule
\end{tabular}
\label{tab:decoding-temp-results}
\end{table}

\begin{figure}[t!]
\centering
\includegraphics[width=0.7\linewidth]{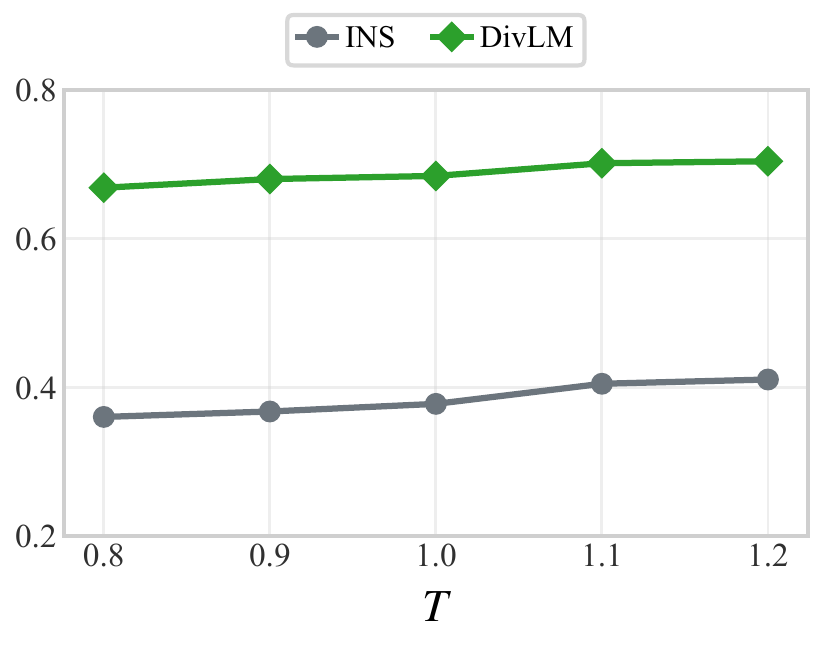}
\caption{Mean diversity reward ($\mathrm{R}_{\text{div}}$) on the test set for \divLM~and INS across decoding temperatures ($T$) on \texttt{Llama-3.1-8B}.}
\label{fig:decoding-temp-div}
\end{figure}

\begin{figure}[t!]
\centering
\includegraphics[width=0.7\linewidth]{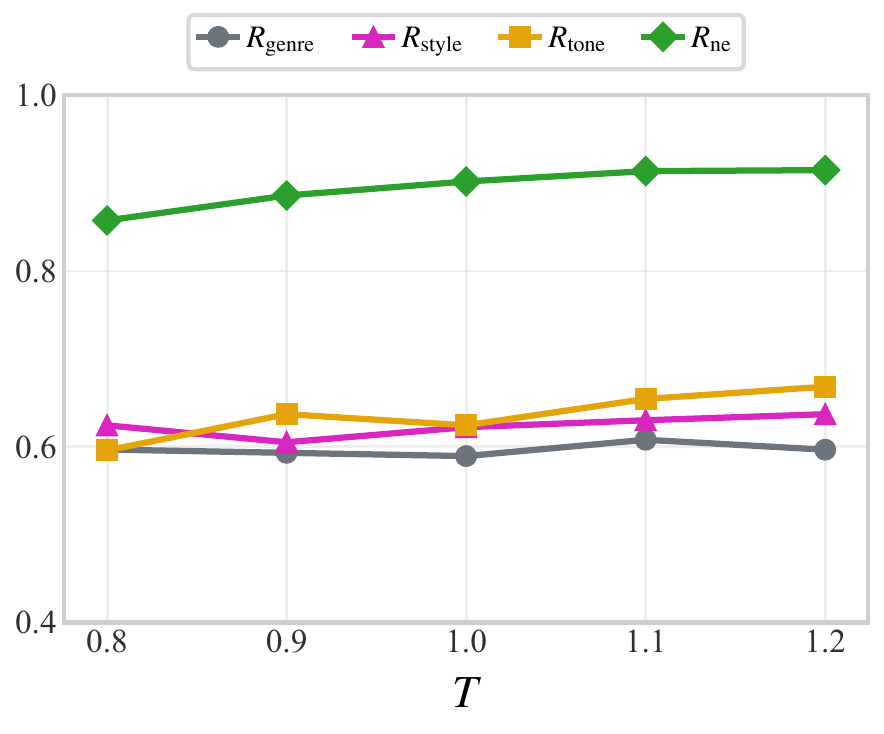}
\caption{\divLM~diversity evaluation results across decoding temperatures ($T$) on \texttt{Llama-3.1-8B}.}
\label{fig:decoding-temp-all}
\end{figure}

\section{Decoding Temperature Analysis}
\label{subsec:decoding-temp}
The decoding temperature $T$ is a hyperparameter that controls output randomness and is commonly used to increase diversity and creativity in LLM outputs~\citep{peeperkorn-etal-2024-temperature, troshin-etal-2025-control}. It modifies the token probability distribution by scaling the logits before softmax normalization~\citep{keskar2019ctrl, renze-2024-effect}. Values of $T > 1$ flatten the distribution, increasing the probability of sampling lower-ranked tokens. To evaluate the performance of our model under different decoding temperatures, we compute the diversity metrics on the test set for $T \in \{0.8, 0.9, 1.0, 1.1, 1.2\}$ and compare the results with the INS model. Higher values of $T$ lead to a similar increase in cosine distance for both the INS and \divLM~models, while the gap between them remains consistent across the evaluated range (Table~\ref{tab:decoding-temp-results}). Similar results are observed for the mean diversity reward (Figure~\ref{fig:decoding-temp-div}). The \divLM~model maintains substantially higher diversity at every temperature. Notably, the INS model at $T=1.2$ does not approach the diversity reward and semantic distance achieved by the \divLM~model at $T=0.8$, which indicates that increasing decoding temperature alone is insufficient to obtain comparable levels of diversity. The complete diversity results for \divLM~at different values of $T$ are depicted in Figure~\ref{fig:decoding-temp-all}. We observe a slight increase in named entity, style, and tone diversity at higher temperatures, while genre diversity remains largely unchanged.

\begin{table}[t]
\centering
\caption{Average cosine distance on the test set for INS and \divLM~ under different group sizes ($G$).}
\small
\setlength{\tabcolsep}{3pt}
\renewcommand{\arraystretch}{0.95}
\begin{tabular}{@{}lrrrrr@{}}
\toprule
Model & $G=5$ & $G=10$ & $G=20$ & $G=50$ & $G=100$ \\
\midrule
INS
& .123 & .125 & .126 & .126 & .126 \\
\divLM
& \textbf{.227} & \textbf{.226} & \textbf{.226}
& \textbf{.226} & \textbf{.226} \\
\bottomrule
\end{tabular}
\label{tab:groupsize-results}
\end{table}

\begin{figure}[b]
\centering
\includegraphics[width=0.7\linewidth]{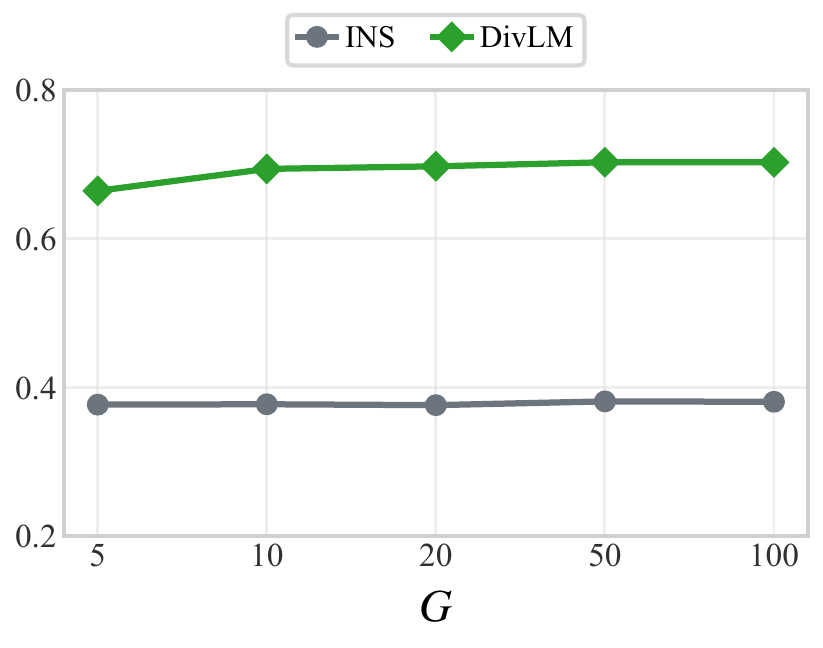}
\caption{Mean diversity reward ($\mathrm{R}_{\text{div}}$) on the test set for \divLM~and INS across group sizes ($G$) on \texttt{Llama-3.1-8B}.}
\label{fig:decoding-group-div}
\end{figure}

\begin{figure}[t]
\centering
\includegraphics[width=0.7\linewidth]{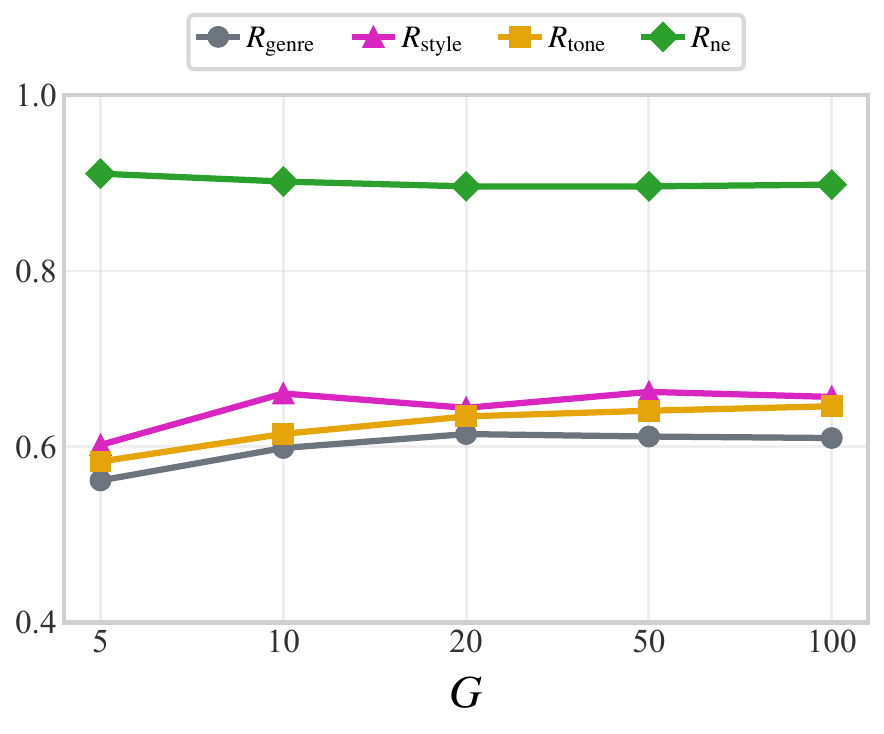}
\caption{\divLM~diversity evaluation results across group sizes ($G$) on \texttt{Llama-3.1-8B}.}
\label{fig:groupsize-all}
\end{figure}

\section{Group Size Analysis}
\label{subsec:group-size}
To evaluate the performance of our model with different group sizes, we compute the diversity metrics on the test set for $G \in \{5, 10, 20, 50, 100\}$ and compare with the INS model. The results show that cosine distance remains approximately constant across different group sizes (Table~\ref{tab:groupsize-results}). The mean diversity reward obtained by \divLM~maintains a consistent gap with INS (Figure~\ref{fig:decoding-group-div}). In addition, we report the complete diversity evaluation results of \divLM~at different values of $G$ (Figure~\ref{fig:groupsize-all}). From $G=5$ to $G=10$, we observe slight increases in genre, tone, and style diversity, along with a slight decrease in named entity diversity. After $G=10$, all metrics stabilize.

\begin{table}[b]
\centering
\small
\setlength{\tabcolsep}{6pt}
\begin{tabular}{lrr}
\toprule
Setting & INS & \divLM \\
\midrule
Decay = 50  & 481.2 & 366.3 \\
Decay = 150 & 481.2 & 410.0 \\
Linear   & 481.2 & 483.6 \\
\bottomrule
\end{tabular}
\caption{Mean output length in words for the INS and  \divLM~models using different length penalties.}
\label{tab:length_stats_decay}
\end{table}

\section{Length Penalty}
\label{sec:length-penalty} 
We observe that RL training may result in shorter model outputs. Although our task does not have an ideal output length, keeping the average output length consistent is necessary for fair comparison, as output length can directly affect evaluation metrics. To mitigate this effect, we apply a length penalty to the GRPO reward with the goal of maintaining output lengths comparable to those of the INS model. An alternative approach would be to compute length-normalized metrics; however, metrics evaluated by LLM-based judges cannot be reliably length-normalized. Thus, we opt to explicitly regulate output length during training rather than normalizing evaluation metrics. We do acknowledge that variation in output length can itself be a form of diversity and may be sometimes desirable.

We first compute the average output length of the INS model, denoted by $L_{\text{avg}}$, and adopt this as the target average during RL. We compare exponential penalties with decay values of 50 and 150 against a linear penalty. In all cases, the length penalty lies in the range $[0, 1]$. For the linear penalty, we calculate the penalty as $\max(0, 1-\frac{L_{o_i}}{L_{\text{avg}}})$, where $L_{o_i}$ is the length of output $o_i$. Table~\ref{tab:length_stats_decay} reports the mean output length under different penalty settings, showing that the linear approach preserves the length relative to the INS model. Accordingly, we adopt this formulation for our length penalty. 

\section{Multi-run Evaluation}
\label{subsec:multirun-eval}
To assess the robustness of our results, we repeat the evaluation across five different seeds (Table~\ref{tab:five-run-results}). Due to computational constraints, we focus on INS, CPT, and \divLM. We also include DDPO, which has the highest diversity score among the baselines. We evaluate quality, genre/tone/style coverage, diversity reward, and cosine distance. For each metric, we report the mean and standard deviation across the five runs. The results show low variability for all models, with standard deviations $\leq$ .009. \divLM~achieves the highest quality, coverage, and diversity reward. These findings reinforce the conclusions of our main evaluation.

\begin{table}[t]
\centering
\caption{Five-run evaluation results for \texttt{Llama-3.1-8B} on the test set. We compare the following model variants: Instruct (INS), CPT, DDPO, and \divLM. We report the mean $\pm$ standard deviation across five runs. Quality corresponds to the average of coherence, prompt adherence, no meta-commentary rate, and completeness. Coverage corresponds to the average of genre, tone, and style coverage. Diversity reward corresponds to the average of $\mu\left(\mathrm{R}_{\text{genre}}\right)$, $\mu\left(\mathrm{R}_{\text{tone}}\right)$, $\mu\left(\mathrm{R}_{\text{style}}\right)$, and $\mu\left(\mathrm{R}_{\text{ne}}\right)$. Greater values indicate better performance.}
\label{tab:five-run-results}
\setlength{\tabcolsep}{4pt}
\renewcommand{\arraystretch}{1.3}
\resizebox{\columnwidth}{!}{%
\begin{tabular}{lcccc}
\toprule
& \multicolumn{4}{c}{\texttt{Llama-3.1-8B}} \\
\cmidrule(lr){2-5}
\bf Metric
& INS
& CPT
& DDPO
& \divLM \\
\midrule
Quality 
    & .922 $\pm$ .004
    & .740 $\pm$ .005
    & .528 $\pm$ .009
    & \bf .991 $\pm$ .002 \\
Coverage
    & .446 $\pm$ .008
    & .555 $\pm$ .003
    & .621 $\pm$ .008
    & \bf .675 $\pm$ .003 \\
Diversity reward
    & .388 $\pm$ .009
    & .567 $\pm$ .003
    & .635 $\pm$ .008
    & \bf .710 $\pm$ .009 \\
Cosine distance
    & .127 $\pm$ .001
    & .172 $\pm$ .001
    & \bf .237 $\pm$ .001
    & .228 $\pm$ .001 \\
\bottomrule
\end{tabular}%
}
\end{table}

\end{document}

%% file: figures/pipeline_tikz.tex

\definecolor{pipeInk}{HTML}{000000}
\definecolor{pipeMuted}{HTML}{667085}
\definecolor{pipeLine}{HTML}{3F4A5A}
\definecolor{pipeCptBack}{HTML}{FFFFFF}
\definecolor{pipeCptEdge}{HTML}{5FAFA7}
\definecolor{pipeRlBack}{HTML}{FFFFFF}
\definecolor{pipeRlEdge}{HTML}{829BC8}
\definecolor{pipeModelFill}{HTML}{FFFFFF}
\definecolor{pipeModelEdge}{HTML}{546274}
\definecolor{pipeDataFill}{HTML}{D6EEEA}
\definecolor{pipeDataEdge}{HTML}{21877F}
\definecolor{pipeRewardFill}{HTML}{E6EEF3}
\definecolor{pipeRewardEdge}{HTML}{315D7C}
\definecolor{pipeVectorFill}{HTML}{D7E4F3}
\definecolor{pipeVectorEdge}{HTML}{38679E}
\definecolor{pipeFinalFill}{HTML}{C9DAEC}
\definecolor{pipeFinalEdge}{HTML}{1F426C}
\definecolor{pipeUpdate}{HTML}{4F6178}
\definecolor{pipeCptHeader}{HTML}{D3EAE6}
\definecolor{pipeRlHeader}{HTML}{D8E2F1}

\tikzset{
  pipePanel/.style={rounded corners=5pt, line width=0.48pt},
  pipeTitle/.style={anchor=west, font=\bfseries\small, text=pipeInk, inner sep=0pt},
  pipeSubTitle/.style={anchor=west, font=\scriptsize, text=pipeMuted, inner sep=0pt},
  pipeBox/.style={
    rectangle,
    rounded corners=3.5pt,
    align=center,
    inner xsep=4pt,
    inner ysep=3pt,
    minimum height=0.64cm,
    line width=0.50pt,
    font=\small,
    text=pipeInk
  },
  pipeModel/.style={pipeBox, draw=pipeModelEdge, fill=pipeModelFill},
  pipeData/.style={pipeBox, draw=pipeDataEdge, fill=pipeDataFill},
  pipeReward/.style={
    rectangle,
    rounded corners=5pt,
    draw=none,
    fill=pipeRewardFill,
    align=left,
    inner xsep=4.5pt,
    inner ysep=4pt,
    line width=0pt,
    font=\small,
    text=pipeInk
  },
  pipeVector/.style={
    pipeBox,
    draw=pipeVectorEdge,
    fill=pipeVectorFill,
    minimum width=1.05cm,
    inner xsep=3.5pt
  },
  pipeFinalVector/.style={pipeVector, draw=pipeFinalEdge, fill=pipeFinalFill, line width=0.9pt},
  pipeOp/.style={
    circle,
    draw=pipeUpdate,
    fill=white,
    minimum size=0.46cm,
    inner sep=0pt,
    line width=0.55pt,
    font=\small
  },
  pipeArrow/.style={-stealth, line width=0.62pt, draw=pipeLine},
  pipeThinArrow/.style={-stealth, line width=0.5pt, draw=pipeLine},
  pipeUpdateArrow/.style={-stealth, dashed, line width=0.50pt, draw=pipeUpdate}
}

\resizebox{\textwidth}{!}{%
\begin{tikzpicture}[x=1cm,y=1cm]

\filldraw[pipePanel, fill=pipeCptBack, draw=pipeCptEdge]
  (-0.25,-3.05) rectangle (8.05,2.85);
\fill[pipeCptHeader, rounded corners=4pt]
  (-0.25,2.98) rectangle (8.05,3.38);
\node[font=\bfseries\small, text=pipeInk, inner sep=0pt]
  at (3.90,3.18) {CPT phase};

\filldraw[pipePanel, fill=pipeRlBack, draw=pipeRlEdge]
  (8.55,-3.05) rectangle (26.60,2.85);
\fill[pipeRlHeader, rounded corners=4pt]
  (8.55,2.98) rectangle (26.60,3.38);
\node[font=\bfseries\small, text=pipeInk, inner sep=0pt]
  at (17.58,3.18) {RL phase};

\node[pipeModel, text width=1.14cm] (instruct) at (0.78,1.18)
  {Instruct\\model};
\node[pipeModel, text width=1.14cm] (base) at (0.78,-1.18)
  {Base\\model};

\node[pipeData, text width=2.12cm] (residual) at (3.55,1.18)
  {Instruction residuals\\[-1pt] from Eq.~\ref{eq:delta}};
\node[pipeData, text width=2.12cm] (corpus) at (3.55,-1.18)
  {Continued pre-training\\on creative writing};

\node[pipeOp] (plus) at (5.55,0) {$+$};
\node[pipeModel, text width=1.05cm] (cpt) at (6.95,0) {CPT\\model};

\draw[pipeArrow] (instruct) -- (residual);
\coordinate (baseFork) at ($(base.east)+(0.07,0)$);
\draw[line width=0.5pt, draw=pipeLine] (base.east) -- (baseFork);
\draw[pipeArrow] (baseFork) -- (corpus.west);
\draw[pipeThinArrow]
  (baseFork) -- ++(0.16,0) |- ($(residual.west)+(0,-0.30)$);
\draw[pipeArrow] (residual.east) -- (plus.north west);
\draw[pipeArrow] (corpus.east) -- (plus.south west);
\draw[pipeArrow] (plus) -- (cpt);

\node[font=\small, align=center, text=pipeInk] (prompt) at (9.70,1.52)
  {Writing\\prompt};
\node[pipeModel, text width=1.44cm] (policy) at (9.70,0)
  {Policy model\\[-1pt]{\scriptsize initialized from CPT}};
\draw[pipeArrow] (prompt) -- (policy);

\node[pipeVector, minimum width=0.86cm, inner xsep=2.5pt] (outputs) at (11.70,0)
  {$o_1$\\[-2pt]$o_2$\\[-3pt]$\vdots$\\[-1pt]$o_G$};
\draw[pipeArrow] (policy) -- (outputs);

\node[pipeModel, text width=1.08cm] (judge) at (13.82,1.05)
  {LLM\\judge};
\node[pipeModel, text width=1.08cm] (rules) at (13.82,-1.05)
  {Rule\\checks};
\draw[pipeArrow] (outputs.east) -- ++(0.34,0) |- (judge.west);
\draw[pipeArrow] (outputs.east) -- ++(0.34,0) |- (rules.west);

\node[pipeReward, text width=2.50cm] (diversity) at (17.15,1.05)
  {\textbf{Diversity}\\[-1pt]
   \textbullet~genre\quad \textbullet~tone\\[-1pt]
   \textbullet~style\quad \textbullet~entity};
\node[pipeReward, text width=2.66cm] (quality) at (17.15,-1.05)
  {\textbf{Quality}\\[-1pt]
   \textbullet~prompt adherence\\[-1pt]
   \textbullet~malformed outputs\\[-1pt]
   \textbullet~meta-commentary\\[-1pt]
   \textbullet~completeness};

\coordinate (divTitle) at ($(diversity.west)+(0,0.35)$);
\coordinate (qualTitle) at ($(quality.west)+(0,0.55)$);
\coordinate (qualMalformed) at ($(quality.west)+(0.16,0)$);
\coordinate (qualMeta) at ($(quality.west)+(0.16,-0.33)$);
\coordinate (qualComplete) at ($(quality.west)+(0.16,-0.66)$);

\coordinate (judgeFork) at ($(judge.east)+(0.18,0)$);
\draw[line width=0.5pt, draw=pipeLine] (judge.east) -- (judgeFork);
\draw[pipeThinArrow] (judgeFork) |- (divTitle);
\draw[pipeThinArrow] (judgeFork) |- (qualTitle);

\coordinate (ruleFork) at ($(rules.east)+(0.18,0)$);
\draw[line width=0.5pt, draw=pipeLine] (rules.east) -- (ruleFork);
\draw[pipeThinArrow] (ruleFork) -- (qualMalformed);
\draw[pipeThinArrow] (ruleFork) |- (qualMeta);
\draw[pipeThinArrow] (ruleFork) |- (qualComplete);

\node[pipeVector, text width=1.24cm] (rdiv) at (20.10,1.05)
  {$\mathrm{R}_{\text{div}}(o_1)$\\[1pt]
   $\mathrm{R}_{\text{div}}(o_2)$\\[-1pt]
   $\vdots$\\[-1pt]
   $\mathrm{R}_{\text{div}}(o_G)$};
\node[pipeVector, text width=1.24cm] (rqual) at (20.10,-1.05)
  {$\mathrm{R}_{\text{qual}}(o_1)$\\[1pt]
   $\mathrm{R}_{\text{qual}}(o_2)$\\[-1pt]
   $\vdots$\\[-1pt]
   $\mathrm{R}_{\text{qual}}(o_G)$};
\draw[pipeArrow] (diversity) -- (rdiv);
\draw[pipeArrow] (quality) -- (rqual);

\node[pipeModel, text width=1.18cm] (combine) at (22.50,0)
  {Combine\\[-1pt](Eq.~\ref{eq:total_reward})};
\draw[pipeArrow] (rdiv.east) -- ++(0.32,0) |- ($(combine.west)+(0,0.18)$);
\draw[pipeArrow] (rqual.east) -- ++(0.32,0) |- ($(combine.west)+(0,-0.18)$);

\node[pipeFinalVector, text width=0.92cm] (reward) at (24.05,0)
  {$\mathrm{R}(o_1)$\\[1pt]
   $\mathrm{R}(o_2)$\\[-1pt]
   $\vdots$\\[-1pt]
   $\mathrm{R}(o_G)$};
\draw[pipeArrow] (combine) -- (reward);

\node[pipeModel, text width=0.92cm] (grpo) at (25.50,0)
  {GRPO\\[-1pt](Eq.~\ref{eq:objective})};
\draw[pipeArrow] (reward) -- (grpo);

\draw[pipeUpdateArrow, rounded corners=3pt]
  (grpo.south) -- (25.50,-2.46) -- (9.70,-2.46)
  node[midway, below, font=\scriptsize, text=pipeUpdate, inner xsep=1pt] {policy update}
  -- (policy.south);

\draw[pipeArrow, draw=pipeCptEdge, dashed]
  (cpt.east) -- node[pos=0.62, above, font=\tiny, text=pipeMuted, inner sep=0.5pt] {init.}
  (policy.west);

\pgfresetboundingbox
\path[use as bounding box] (-0.25,-3.05) rectangle (26.60,3.38);
\end{tikzpicture}%
}

%% file: tables/main-perf.tex
\begin{table}[t]
\centering
\caption{Comparison of the original Instruct (INS) LLM variants with CPT-tuned and \divLM~(shortened to \divLMshort) models across quality and diversity metrics ranging in $[0,1]$. For \texttt{Llama-3.1-8B}, we include DQO and DDPO. To allow the calculation of meaningful averages, we report the no meta-commentary rate. 
Greater ($\uparrow$) values indicate better performance.}
\label{tab:final-results}
\setlength{\tabcolsep}{2pt}
\resizebox{\columnwidth}{!}{%
\begin{tabular}{clllllllll}
\toprule
& & \multicolumn{5}{c}{\texttt{Llama-3.1-8B}}
& \multicolumn{3}{c}{\texttt{Gemma-2-9B}} \\
\cmidrule(lr){3-7}
\cmidrule(lr){8-10}
& \bf Metric
& \multicolumn{1}{c}{INS}
& \multicolumn{1}{c}{CPT}
& \multicolumn{1}{c}{DQO}
& \multicolumn{1}{c}{DDPO}
& \multicolumn{1}{c}{\divLMshort}
& \multicolumn{1}{c}{INS}
& \multicolumn{1}{c}{CPT}
& \multicolumn{1}{c}{\divLMshort}\\
\midrule
\multirow{5}{*}[-1ex]{\rotatebox{90}{\bf Quality}} &
Coherence
    & .997 & .999 & .999 & .886 & .997
    & 1    & .998 & .995 \\
& Prompt adherence
    & .997 & .995 & .999 & .834 & .981
    & .999 & .998 & .990 \\
& No meta-comments
    & .968 & .865 & .952 & .399 & .998
    & .983 & .956 & .998 \\
& Completeness
    & .726 & .111 & .983 & .017 & .985
    & .633 & .128 & .979 \\
\cmidrule(lr){3-10}
& \sc{Avg}
    & .922 & .742 & .983 & .534 & \bf .990
    & .904 & .770 & \bf .990 \\
\midrule
\multirow{10}{*}[-2ex]{\rotatebox{90}{\bf Diversity}} &
Genre coverage
    & .210 & .270 & .226 & .341 & .382
    & .224 & .265 & .312 \\
& Tone coverage
    & .603 & .717 & .572 & .853 & .832
    & .657 & .761 & .741 \\
& Style coverage
    & .542 & .684 & .542 & .657 & .801
    & .542 & .623 & .811 \\
\cmidrule(lr){3-10}
& \sc{Avg}
    & .452 & .557 & .447 & .617 & \bf .672
    & .474 & .550 & \bf .621    \\
\cmidrule(lr){2-10}
& $\mu\left(\mathrm{R}_{\text{genre}}\right)$
    & .293 & .413 & .354 & .567 & .606
    & .322 & .417 & .506 \\
& $\mu\left(\mathrm{R}_{\text{tone}}\right)$
    & .345 & .532 & .336 & .698 & .667
    & .429 & .567 & .542 \\
& $\mu\left(\mathrm{R}_{\text{style}}\right)$
    & .291 & .546 & .294 & .435 & .644
    & .296 & .454 & .593 \\
& $\mu\left(\mathrm{R}_{\text{ne}}\right)$
    & .581 & .788 & .535 & .852 & .902
    & .676 & .794 & .806 \\
\cmidrule(lr){3-10}
& \sc{Avg}
    & .378 & .570 & .380 & .638 & \bf .705
    & .431 & .558 & \bf .612 \\
\cmidrule(lr){2-10}
& Cosine distance
    & .125 & .174 & .135 & \bf .236 & .226
    & .124 & .152 & \bf .179 \\
\bottomrule
\end{tabular}%
}
\end{table}